\documentclass{article}

\PassOptionsToPackage{numbers, compress}{natbib}
\usepackage[preprint]{neurips_data_2021}

\usepackage[utf8]{inputenc} 
\usepackage[T1]{fontenc}    
\usepackage{hyperref}       
\usepackage{url}            
\usepackage{natbib}
\usepackage{booktabs}       
\usepackage{amsfonts}       
\usepackage{nicefrac}       
\usepackage{microtype}      
\usepackage{xcolor}         
\usepackage{amsmath}
\usepackage{graphicx}
\usepackage[scriptsize]{subfigure}
\usepackage{enumerate}
\usepackage{booktabs}
\usepackage{multirow}
\usepackage{multicol}
\usepackage{listings}

\usepackage{listings}
\usepackage{xcolor}

\usepackage{verbatim}
\usepackage{tikz}
\usetikzlibrary{positioning,quotes,calc,fit,shapes,shadows,arrows,decorations.pathreplacing}
\usepackage[font=scriptsize]{caption}
\newcommand{\norm}[1]{\left\lVert#1\right\rVert}
\definecolor{bp-color}{rgb}{1.0, 0.0, 0.16}
\definecolor{fa-color}{rgb}{1.0, 0.7281399046104929, 0.0}
\definecolor{dfa-color}{rgb}{0.36036036036036034, 1.0, 0.0}
\definecolor{uSF-color}{rgb}{0.0, 1.0, 0.5481762597512125}
\definecolor{brSF-color}{rgb}{0.0, 0.5615942028985503, 1.0}
\definecolor{frSF-color}{rgb}{0.35485933503836337, 0.0, 1.0}

\definecolor{codegreen}{rgb}{0,0.6,0}
\definecolor{codegray}{rgb}{0.5,0.5,0.5}
\definecolor{codepurple}{rgb}{0.58,0,0.82}
\definecolor{backcolour}{rgb}{0.95,0.95,0.92}
\definecolor{github-link}{RGB}{0,0,139}
\lstdefinestyle{mystyle}{
    backgroundcolor=\color{backcolour},   
    commentstyle=\color{codegreen},
    keywordstyle=\color{magenta},
    numberstyle=\tiny\color{codegray},
    stringstyle=\color{codepurple},
    basicstyle=\ttfamily\footnotesize,
    breakatwhitespace=false,         
    breaklines=true,                 
    captionpos=b,                    
    keepspaces=true,                 
    numbers=left,                    
    numbersep=5pt,                  
    showspaces=false,                
    showstringspaces=false,
    showtabs=false,                  
    tabsize=2
}

\newcommand\YAMLcolonstyle{\color{red}\mdseries}
\newcommand\YAMLkeystyle{\color{black}\bfseries}
\newcommand\YAMLvaluestyle{\color{blue}\mdseries}

\makeatletter

\newcommand\language@yaml{yaml}

\expandafter\expandafter\expandafter\lstdefinelanguage
\expandafter{\language@yaml}
{
  keywords={true,false,null,y,n},
  keywordstyle=\color{darkgray}\bfseries,
  basicstyle=\YAMLkeystyle,                                 
  sensitive=false,
  comment=[l]{\#},
  morecomment=[s]{/*}{*/},
  commentstyle=\color{purple}\ttfamily,
  stringstyle=\YAMLvaluestyle\ttfamily,
  moredelim=[l][\color{orange}]{\&},
  moredelim=[l][\color{magenta}]{*},
  moredelim=**[il][\YAMLcolonstyle{:}\YAMLvaluestyle]{:},   
  morestring=[b]',
  morestring=[b]",
  literate =    {---}{{\ProcessThreeDashes}}3
                {>}{{\textcolor{red}\textgreater}}1     
                {|}{{\textcolor{red}\textbar}}1 
                {\ -\ }{{\mdseries\ -\ }}3,
}

\lst@AddToHook{EveryLine}{\ifx\lst@language\language@yaml\YAMLkeystyle\fi}
\makeatother

\newcommand\ProcessThreeDashes{\llap{\color{cyan}\mdseries-{-}-}}

\title{Benchmarking the Accuracy and Robustness of Feedback Alignment Algorithms}

\author{%
  Albert Jiménez Sanfiz\thanks{Corresponding author} \\
  AIP Labs\\
  \texttt{albert@aip.ai} \\
  \And
  Mohamed Akrout \\
  AIP Labs\\
  \texttt{mohamed@aip.ai} \\
}

\begin{document}

\maketitle

\begin{abstract}
Backpropagation is the default algorithm for training deep neural networks due to its simplicity, efficiency and high convergence rate. However, its requirements make it impossible to be implemented in a human brain. In recent years, more biologically plausible learning methods have been proposed. Some of these methods can match backpropagation accuracy, and simultaneously provide other extra benefits such as faster training on specialized hardware (e.g., ASICs) or higher robustness against adversarial attacks. While the interest in the field is growing, there is a necessity for open-source libraries and toolkits to foster research and benchmark algorithms. In this paper, we present \textit{BioTorch}, a software framework to create, train, and benchmark biologically motivated neural networks. In addition, we investigate the performance of several feedback alignment methods proposed in the literature, thereby unveiling the importance of the forward and backward weight initialization and optimizer choice. Finally, we provide a novel robustness study of these methods against state-of-the-art white and black-box adversarial attacks.

\end{abstract}

\section{Introduction}\label{sec:introduction}
Backpropagation (BP) ~\cite{rumelhart1986learning} is the dominant algorithm to compute the gradients of a learning system with respect to a predefined loss function. Its simplicity, efficiency, and high accuracy and convergence rates, make it the de facto algorithm to train neural networks. However, there is evidence that such an algorithm could not be biologically implemented by the human brain. One of the main reasons is the need of BP to have symmetric forward and backward synaptic weights (i.e., the weight transport or the weight symmetry problem)~\cite{rumelhart1986learning, grossberg1987competitive, stork1989backpropagation, crick1989recent, chinta2012adaptive}. Since synapses are unidirectional in the brain, feedforward and feedback connections must be physically distinct.

To overcome the aforementioned
limitation, recent studies in learning algorithms have focused on the intersection between neuroscience and machine learning by studying more biologically-plausible algorithms. One of the main family of methods proposed is based on neural activity differences to encode errors (NGRAD)~\cite{lillicrap2020backpropagation}, namely, target propagation (TP) and its variants~\cite{lee2015difference, bengio2014auto, hinton2007backpropagation, le1986learning, bartunov2018assessing, bengio2020deriving}. The other family is known as alignment methods, which employ distinct forward and feedback synaptic weights, where the latter are randomized~\cite{lillicrap2014random, lillicrap2016random, nokland2016direct}, thereby providing noisy and synthetic gradient training schemes. Neural networks trained with alignment methods learn to adapt the forward synaptic weights so the alignment between both matrices increase, making learning possible. Therefore, random feedback weights can transmit useful teaching signals to neurons along the network depth~\cite{lillicrap2014random}. The scope of this work will be focused on feedback alignment (FA) methods.

Early studies~\cite{lillicrap2014random, nokland2016direct}, have shown that shallow networks using FA and direct FA (DFA) methods were able to learn good feature representations of complete simple tasks such digit recognition in MNIST. Deeper architectures were trained in~\cite{bartunov2018assessing} with FA and TP methods showing their inability to learn in complex visual tasks such as ImageNet~\cite{deng2009imagenet}. The forward-backward weight asymmetry of these methods can lead to vanishing or exploding gradients because the backward weighst remain fixed during the training. Several works~\cite{liao2016important, xiao2018biologically, moskovitz2018feedback, akrout2019deep, kunin2020two} have shown, however, that increasing the alignment of the feedforward and backward weights, alone or in combination with random feedback weight matrices, can help prevent that behaviour. Some works aim to control the magnitudes of the weights at training time. For example, in~\cite{moskovitz2018feedback}, the authors experiment constraining the magnitude of the forward weights after every training iteration scaling them by the initial weights $L_2$ norm. In~\cite{liao2016important}, the authors transport the sign of the weight matrix, and propose Batch Manhattan (BM) as a parameter update rule, where the gradient magnitudes are discarded and the updates are proportional to the sign. Another technique proven to be useful is careful initialization of the forward and backward weight matrices. Using variance preserving methods like Xavier~\cite{glorot2010understanding} helps ensure the variance of the activations are the same across every layer, preventing the forward and backward signals from exploding or vanishing. These lastest works reach or surpass BP metrics on MNIST or CIFAR10~\cite{krizhevsky2009learning} and scale to more difficult tasks, obtaining a close performance to BP on ImageNet. 

In addition to providing synthetic gradient training schemes, it has been shown that alignment methods can increase the robustness of deep learning models against adversarial attacks~\cite{akrout2019adversarial}. Adversarial vulnerability is an intrinsic property of all classifiers~\cite{goodfellow2014explaining} and state-of-the-art defense techniques rely on robust pre-trained models and expensive adversarial training with Projected Gradient Descent (PGD)~\cite{madry2017towards}. It was demonstrated in~\cite{cappelli2021ropust} that fine-tuning pre-trained models using the DFA method increases the robustness when training with a photonic co-processor for physical parameter obfuscation~\cite{cappelli2021adversarial}. Finally, the additional interest for feedback alignment is also driven by their ability to allow forward and feedback weights to live locally in application-specific integrated circuits (ASICs), which ultimately allows for time and energy savings~\cite{chen2016eyeriss, hyoukjun2018maestro, launay2020hardware}.

While the research and interest on feedback alignment methods is growing, we noticed a lack of open-source libraries and public algorithmic benchmarks. In this work, we provide \textit{BioTorch}, a modular and extensible framework with the implementations of five alignment methods: FA, DFA, and sign concordant variants (uSF, brSF, frSF), whose code is available at \href{https://github.com/jsalbert/biotorch}{\textcolor{github-link}{\texttt{https://github.com/jsalbert/biotorch}}}.

Our contributions can be summarized as follows:

\begin{enumerate}[i)]
  \item We provide the first open-source software framework that allows for the creation of custom neural networks with feedback alignment layers, the automatic conversion of model architectures, reproducible training, evaluation, and benchmarking (in both accuracy and robustness against adversarial attacks) via configurable experiments.
  \item We carried out a benchmark study of the performance of feedback alignment algorithms on different datasets. First, we benchmarked a shallow network on MNIST and Fashion-MNIST. Second, we increased the network’s depth as we investigated the impact of different optimizers on the performance of all alignment methods on CIFAR-10. Finally, we benchmarked all methods by training a ResNet-18 on ImageNet.
  \item We performed a novel robustness study of feedback alignment algorithms against a set of state-of-the-art white and black-box adversarial attacks.
\end{enumerate}

\section{Feedback Alignment Methods and Their Potential Adversarial Robustness}\label{sec:alignment-methods}
Given a labeled dataset $\cup_k\{(\mathbf{x}^{(k)}, \mathbf{y}^{(k)})\}$ and cost function $\mathcal{L}$, the training of a neural network of $N$ layers alternates between forward passes, to perform inference and compute an error signal, and backward passes, to send the error signal back and update the weights. The output of the $i$th layer, $\mathbf{y}_i$, is computed by forwarding the output signal of the $(i-1)$th signal, $\mathbf{y}_{i-1}$, from layer $(i-1)$ as follows
\begin{equation}
\label{eq:forward-pass}
\mathbf{y}_{i} = \mathbf{\phi}(\mathbf{z}_{i}),~\textrm{with} ~~\mathbf{z}_{i}=\mathbf{W}_{i} \;\mathbf{y}_{i-1} + \mathbf{b}_{i}.
\end{equation}
The backpropagation equation~\cite{rumelhart1986learning} for the weight updates is given by:
\begin{equation}\label{eq:BP}
    \delta \mathbf{W}_{i} = -\left(\left(\mathbf{W}_{i+1}^\top\,\delta\mathbf{z}_{i+1}\right)\,\odot\,\phi^\prime(\mathbf{z}_{i})\right)\mathbf{y}_{i-1}^\top,~\textrm{with} ~~ \delta\mathbf{z}_{i+1} = \frac{\partial\mathcal{L}}{\partial\mathbf{z}_{i+1}}.
\end{equation}

\subsection{Alignment Methods}\label{sec:alignment-methods_subsect}
\begin{figure}[h!]
\centering
\subfigure[BP]{
\includegraphics[scale=0.4]{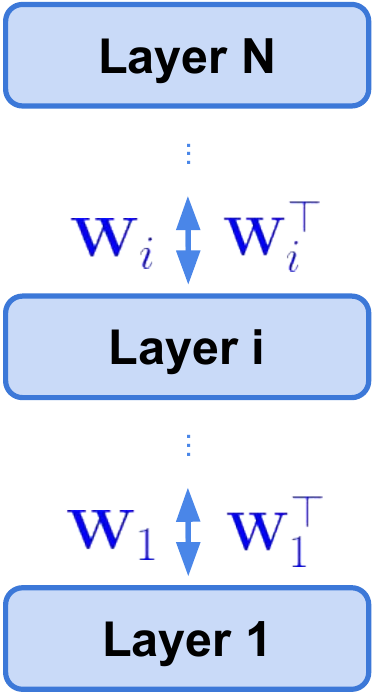}
}\hspace{7mm}
\subfigure[FA]{
\includegraphics[scale=0.4]{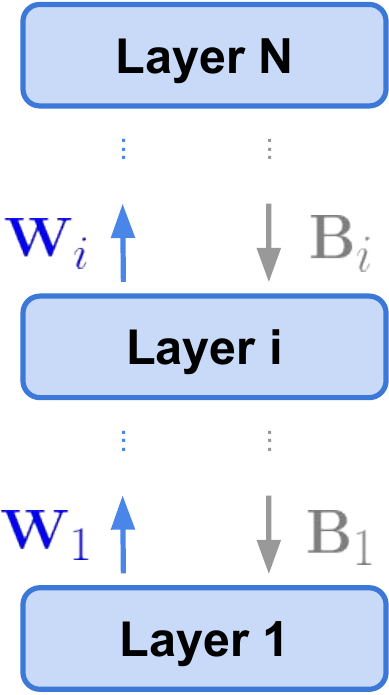}
}\hspace{7mm}
\subfigure[DFA]{
\includegraphics[scale=0.4]{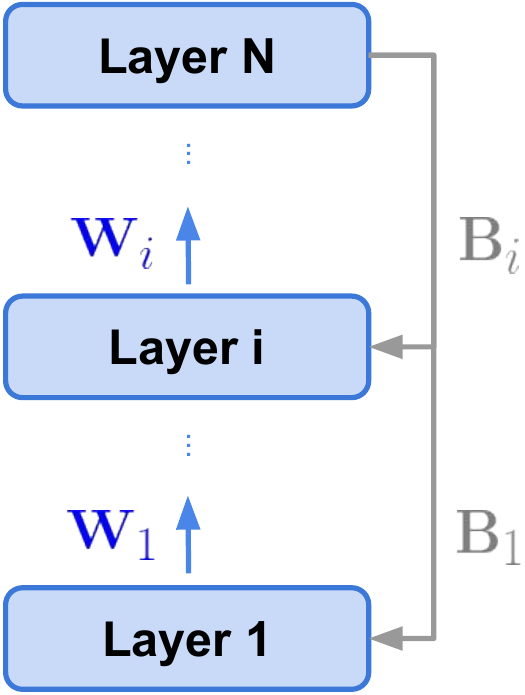}
}
\caption{Schematic overview of different error transportation methods. Blue arrows indicate paths involving the forward pass weight matrix, $\mathbf{W}$, and gray arrows indicate backward paths involving a fixed weight matrix $\mathbf{B}$.}
\label{fig:alignment-methods}
\end{figure}
\textbf{Feedback Alignment (FA)}~~ The weight update (\ref{eq:BP}) of layer $i$ requires the knowledge of the matrix $\mathbf{W}_{i+1}^\top$ and is biologically implausible because it requires that neurons send to each other large numbers of synaptic weights (i.e. weight transport). This fact has motivated the substitution of the matrix $\mathbf{W}_{i+1}^\top$ with a random synaptic weights matrix $\mathbf{B}_{i+1}$ to account for a separate backward pass needed to avoid the weight transport problem~\cite{lillicrap2016random}. Therefore, the weight update becomes:
\begin{equation}\label{eq:FA}
    \delta \mathbf{W}_{i} = -\left(\left(\mathbf{B}_{i+1}\,\delta\mathbf{z}_{i+1}\right)\,\odot\,\phi^\prime(\mathbf{z}_{i})\right)\mathbf{y}_{i-1}^\top,~\textrm{with} ~~ \delta\mathbf{z}_{i+1} = \frac{\partial\mathcal{L}}{\partial\mathbf{z}_{i+1}}.
\end{equation}

\textbf{Direct Feedback Alignment (DFA)}~~While the weight update in feedback alignment is computed recursively across layers, it is possible to project the error propagation by directly backwarding the derivative of the loss at the last layer, $\delta\mathbf{z}_N$, to all layers~\cite{nokland2016direct}. This results in the following update:
\begin{equation}\label{eq:DFA}
    \delta \mathbf{W}_{i} = -\left(\left(\mathbf{B}_{i}\,\delta\mathbf{z}_{N}\right)\,\odot\,\phi^\prime(\mathbf{z}_{i})\right)\mathbf{y}_{i-1}^\top,~\textrm{with} ~~ \delta\mathbf{z}_{N} = \frac{\partial\mathcal{L}}{\partial\mathbf{z}_{N}}.
\end{equation}
where $\mathbf{B}_{i}$ is a fixed random matrix of appropriate shape (i.e., input dimension of layer $i$ $\times$ output dimension of the last layer $N$).

\textbf{Uniform Sign-concordant Feedbacks (uSF)}~~ This method transports the sign of the forward matrices by assuming synaptic weights with unit magnitudes, i.e., $\mathbf{B}_{i} = \textrm{sign}(\mathbf{W}_{i}^\top)~\forall i$.


\textbf{Batchwise Random Magnitude Sign-concordant Feedbacks (brSF)}~~ Instead of assuming a unit magnitude for the synaptic weights of the $i$th layer, this method re-draws their magnitude $\left| \mathbf{R}_i \right|$ after each update such that $\mathbf{B}_i = \left | \mathbf{R}_i \right | \odot \textrm{sign}(\mathbf{W}_i^\top)~\forall i$.


\textbf{Fixed Random Magnitude Sign-concordant Feedbacks (frSF)} This is a variation of the brSF method where the magnitude of weights $\left| \mathbf{R}_i \right|$ is not redrawn after each update, but rather fixed and initialized at the start of the training.

Note that for the uSF method to converge, it is common to scale the sign matrix by the initialization rule applied to create the weight matrix~\cite{liao2016important, moskovitz2018feedback}. For the brSF and frSF methods, we highlight the application of the absolute value of the feedback weights matrix, as opposed to~\cite{liao2016important}, to transport the sign correctly and make convergence possible without the need for an update parameter rule like Batch Manhattan. 


\subsection{Adversarial Robustness}

White-box adversarial attacks are very sensitive to the quality of gradients as they rely on them to perturb the input and fool the model. The noise introduced in those gradients when they are computed using the methods described in Section~\ref{sec:alignment-methods_subsect}, could decrease the effectiveness of such attacks on models trained with feedback alignment.

To investigate the robustness of deep networks trained with alignment methods, it is important to select attacks based on different principles. Black-box attacks have recently become more popular as their strategies are designed to work without requiring access to the target model. We chose to benchmark on four state-of-the-art white-box attacks: fast gradient sign method (FGSM)~\cite{goodfellow2014explaining}, projected gradient descent (PGD)~\cite{madry2017towards}, averaged PGD (APGD)~\cite{zimmermann2019comment}, and PGD used in the TRADES training (TPGD)~\cite{zhang2019theoretically}; and two state-of-the-art black-box attacks:  Few-Pixel~\cite{su2019one} and Square~\cite{andriushchenko2020square}. See the description of these attacks in Appendix~\ref{appendix:adversarial-attacks}.


\section{BioTorch}\label{sec:biotorch}
\textit{BioTorch} is an open-source Python library specializing in biologically plausible learning algorithms that runs \textit{PyTorch}~\cite{paszke2019PyTorch} in the backend for its efficiency and auto-differentiation routines. Its design focuses on the principles of research reproducibility~\cite{pineau2020improving} and integration with \textit{PyTorch} training and testing patterns. \textit{BioTorch} holds implementations of all the feedback alignment methods mentioned in Section~\ref{sec:alignment-methods}. At the moment, it is focused on computer vision tasks, as these have been the most explored in the related literature. 

\subsection{BioTorch Interface}
\textit{BioTorch}'s intuitive interface, allows the user to create neural network models with feedback alignment methods in an effortless manner. Its main features are highlighted below.

\textbf{Model importing:} \textit{BioTorch} comes with a list of predefined 25 state-of-the-art architectures, such as ResNet-50, DenseNet-161 and MNasNet. To create one of these models, it is enough to import the model from the path \texttt{biotorch.models.<method>}, e.g.:
\begin{lstlisting}[language=Python, basicstyle=\small,numbers=none, label={ls:biotorch_model_creation}.]
# Create a ResNet-18 model with FA layers
from biotorch.models.fa import resnet18
model = resnet18()
\end{lstlisting}

\textbf{Custom model creation using alignment layers:} By importing individual \textit{Linear} or \textit{Convolutional} layers implementing alignment methods, the architecture of a neural network model can be defined as follows:
\begin{lstlisting}[language=Python, basicstyle=\small,numbers=none, label={ls:biotorch_custom_model_creation}.]
# Import a convolutional and linear layer using the uSF alignment method
import torch.nn.functional as F
from biotorch.layers.usf import Conv2d, Linear

# Define the custom model using the imported layers
class Model(nn.Module):
  def __init__(self):
    super(Model, self).__init__()
    self.conv1 = Conv2d(in_channels=3, out_channels=128, kernel_size=3)
    self.fc = Linear(in_features=128, out_features=10)

  def forward(self, x):
    out = F.relu(self.conv1(x))
    out = F.avg_pool2d(out, out.size()[3])
    return self.fc(out)

model = Model()
\end{lstlisting}

\textbf{Automatic conversion of an existing model architecture:} It is possible to use the \texttt{BioModule} wrapper of \textit{BioTorch} to automatically convert an existing \textit{PyTorch} model to use any of the alignment methods aforementioned.
\begin{lstlisting}[language=Python, basicstyle=\small,numbers=none, label={ls:biotorch_automatic_conversion_model}.]
from torchvision.models import alexnet
from biotorch.module.biomodule import BioModule

# Convert Alex-Net from using BP to the frSF alignment method
model = BioModule(module=alexnet(), mode='frsf')
\end{lstlisting}

\subsection{Configurable and Reproducible Experiments}

The current version of \textit{BioTorch} allows the user to benchmark different alignment algorithms on a selection of datasets. All the datasets supported are downloaded using the \textit{Torchvision} library, except for ImageNet (\textit{ILSVRC2012}), which is pulled from the official source. 

Diligent measures were taken to ensure correct experiment reproducibility. Experiments are run via configuration files that contain all the training hyperparameters, a dataset, and selected metrics. We include the choice of a fixed seed to set deterministic mathematical and CUDA operations~\cite{reproducibilityPyTorch} to guarantee that the experiments can be replicated. The best performing model on the validation set, the original configuration file, test results, and  \textit{TensorBoard} logs for all the metrics are saved at the end of each experiment. All experiments in Section~\ref{sec:experiments} were run with the following command using the appropriate \texttt{.yaml} configuration file:
\begin{lstlisting}[language=bash,numbers=none,label={biotorch_bash_call}]
#!/bin/bash
python benchmark.py --config benchmark_config_example.yaml
\end{lstlisting}


We included an explained example of a configuration file in the Appendix~\ref{appendix:conf-file} and we refer the reader to the code repository where interactive Colaboratory notebooks and a complete description of the configuration file are provided.

\subsection{Trackable Alignment Metrics}
To observe and measure the behaviour of the algorithms during training, \textit{BioTorch} includes two key metrics of alignment methods.

\textbf{The backward-forward weight norm ratio}~~During the training of networks with feedback alignment methods, the magnitude of the forward weight matrices $\norm{\mathbf{W}^{\top}_i}_2$ changes while the magnitude of feedback weight matrices $\norm{\mathbf{B}_i}_2$ is fixed. This weight update asymmetry can lead to network convergence issues. Indeed, as observed in~\cite{moskovitz2018feedback}, the ratio between the norms of the backward and forward weights, i.e., $\norm{\mathbf{B}_i}_2 ~/~ \norm{\mathbf{W}^{\top}_i}_2$, determine if the trained networks are at risk of experiencing vanishing/exploding gradients when BP does not, even while using the same training hyperparameters.

\textbf{The backward-forward weight alignment}~~For a given layer $i$, the angle between every pair of transposed forward and backward weight matrices, namely, ($\mathbf{W}^\top_i, \mathbf{B}_i$), can be computed to measure how close the considered alignment method is to the case with symmetric weights. Angles close to $0$ denote higher alignment and thus approaching the alignment of BP.

\section{Experiments}\label{sec:experiments}
In accordance with the techniques proposed in past works, we kept $\mathbf{W}$ and $\mathbf{B}$ in the same magnitude scale to improve training in asymmetric conditions. In particular, we experimented constraining the forward weights, discarding the magnitude of the gradients using only their sign or clipping them, and providing a careful initialization of parameters. At the end, the below two choices were the most impactful. 

\textbf{Variance preserving initialization} ~~We initialized our layers weight matrices, $\mathbf{W}$ and $\mathbf{B}$, using Xavier Uniform initialization~\cite{glorot2010understanding}. For the uSF method, we scaled the sign of the weights by the standard deviation of Xavier initialization. For the brSF and frSF methods, we used the absolute value of the random feedback weight matrix to prevent the change of the sign, in contrast with the implementation of~\cite{liao2016important}. Our experiments with these initializations were successful to train all the networks, independently of the network depth.

\textbf{The choice of Adam optimizer}~~ A review of empirical experiments of prior studies, reveals a variety of different optimizer choices. More specifically,~\cite{nokland2016direct} chose RMSprop~\cite{hinton2012neural},~\cite{bartunov2018assessing, moskovitz2018feedback} made use of Adam~\cite{kingma2014adam}, and~\cite{liao2016important, xiao2018biologically, akrout2019deep} used SGD. Our experiments show a significant gap in the classification accuracy of the models depending on the choice of the optimizer, specially for FA and DFA, even after tuning their respective learning rate. For this reason, we show the results of our experiments for both SGD and Adam optimizers.

\subsection{Accuracy Benchmark}\label{subsection:accuracy-benchmark}
\textbf{MNIST \& Fashion-MNIST} ~~We start our empirical study by benchmarking all the alignment methods for LeNet~\cite{lecun1998gradient} on the MNIST and Fashion-MNIST datasets. The networks were trained with the SGD optimizer setting a momentum of $0.9$, and weight decay of $10^{-3}$. We trained for $100$ epochs, decreasing the initial learning rate by a factor of $2$ at the $50th$ and $75th$ epoch. The rest of network hyperparameters can be found in the Appendix (Section~\ref{sec:experiment_details}).

The results of our implementation (Table~\ref{tab:mnisterror}) are close to those reported in~\cite{moskovitz2018feedback} for similar benchmarking conditions. We observe that the performances of FA and DFA are close to BP, and that sign concordant methods match BP performance on MNIST. If we increase the dataset difficulty (Fashion-MNIST), the performance gap between backpropagation and the other methods, specially with the ones not using sign concordant feedback, also increases. 
\begin{table}[h!]
\centering
\begin{tabular}{@{}ccc@{}}
\cmidrule(l){2-3}
        & \multicolumn{2}{c}{Dataset}                                   \\ \cmidrule(l){2-3} 
 & \multicolumn{1}{l}{MNIST} & \multicolumn{1}{l}{Fashion MNIST} \\ \midrule
BP                            & 0.91                      & 9.20                     \\ \midrule
FA                            & 1.7                      & 13.06                             \\ \midrule
DFA                           & 1.61                      & 12.81 
            \\ \midrule
uSF                           & 0.94                      & 9.69                              \\ \midrule
brSF                          & 0.91             & 10.02                              \\ \midrule
frSF                          & 0.97                      & 9.61                
                            \\ \bottomrule
\end{tabular}
\vspace{0.1cm}
\caption{Top-1 test error (\%) of LeNet on the MNIST and Fashion-MNIST datasets.}
\label{tab:mnisterror}
\end{table}

\textbf{CIFAR-10 \& ImageNet} ~~What happens when we scale the application of alignment methods to deeper architectures and more challenging tasks? To answer this question, we benchmark three different networks on CIFAR-10. We start with a modified 5-layer LeNet network, and continue increasing the depth with a ResNet-20, followed by a ResNet-56~\cite{he2016deep}. We applied the same dataset split as in~\cite{he2016deep} where a 5K images from the training set were sampled and used as validation test, while the evaluation is performed on the unseen test set. All the training details and network hyperparameters can be found in the Appendix Section~\ref{sec:experiment_details}. 

The use of an adaptive parameter independent optimizer as Adam outperformed SGD in all the network configurations for the FA and DFA methods, as shown in Table~\ref{tab:cifar10results}. These methods are the ones where the asymmetry in the backward pass is larger.

\begin{table}[h!]
\centering
\begin{tabular}{@{}ccccccc@{}}
\cmidrule(l){2-7}
 &\multicolumn{2}{c}{LeNet} &\multicolumn{2}{c}{ResNet-20} &\multicolumn{2}{c}{ResNet-56}\\
\cmidrule(l){2-3} \cmidrule(l){4-5} \cmidrule(l){6-7} 
 & SGD & Adam & SGD & Adam & SGD & Adam \\ \midrule
BP     & 14.23 & 15.92        & 8.63      & 10.01            & 8.3             & 7.83             \\ \midrule
FA     & 46.69 & 40.67        & 32.16     & 29.59            & 34.88           & 29.23            \\ \midrule
DFA    & 54.21 & 37.59        & 45.94     & 32.16            & 38.01           & 32.02            \\ \midrule
uSF    & 16.22 & 16.34        & 10.05     & 10.59            & 8.2             & 9.19             \\ \midrule
brSF   & 16.02 & 17.08        & 11.02     & 11.08            & 8.69            & 10.13            \\ \midrule
frSF   & 16.86 & 16.83        & 11.2      & 11.22            & 9.49            & 10.02            \\ \bottomrule
\end{tabular}
\vspace{0.1cm}
\caption{Top-1 test error (\%) on the CIFAR-10 dataset. Three networks ranging from shallower to deeper are benchmarked using BP and alignment methods.}
\label{tab:cifar10results}
\end{table}

\begin{figure}[h!]
    \centering
    \subfigure[DFA (SGD)]{\includegraphics[width=0.23\textwidth]{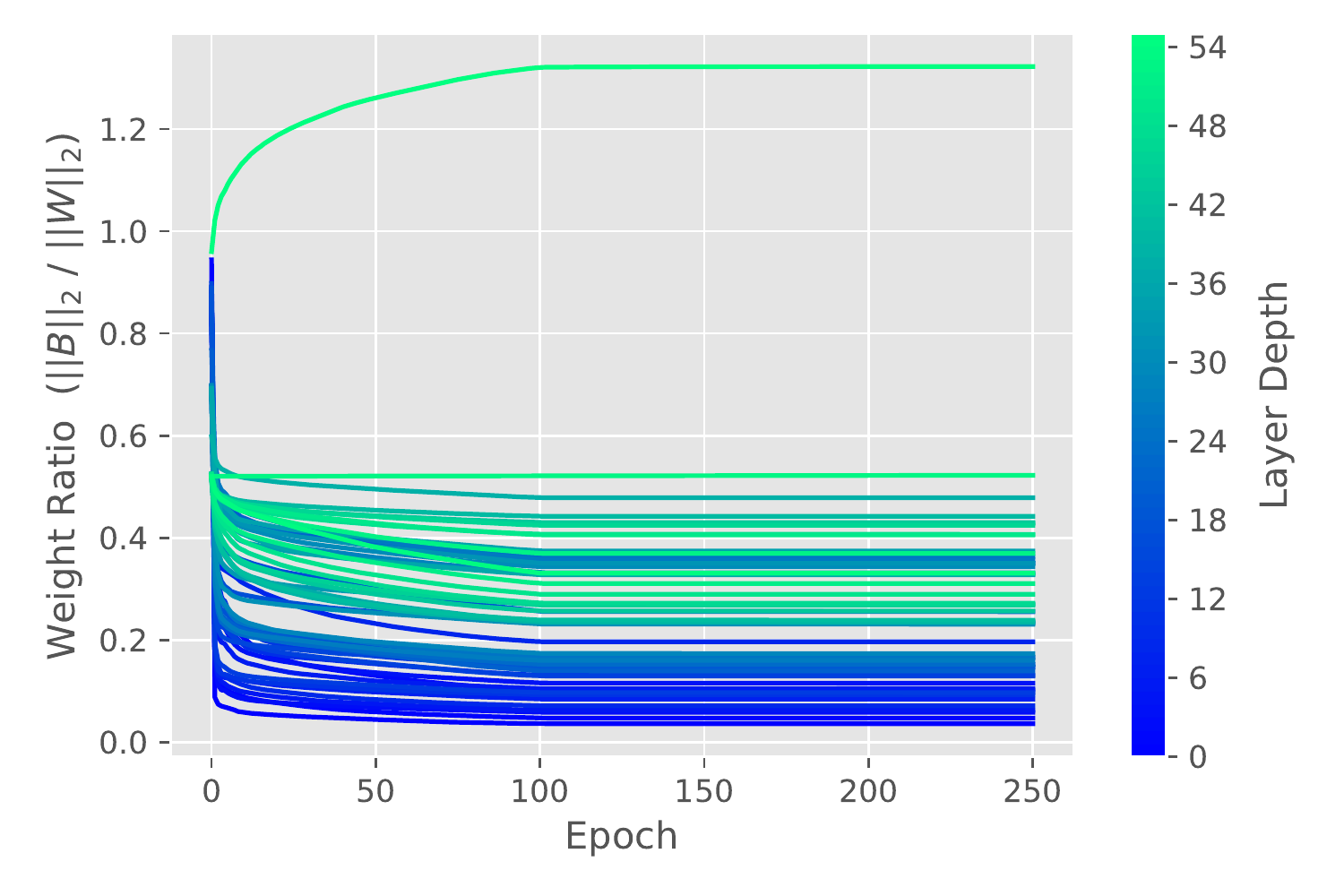}}
    \subfigure[DFA (Adam)]{\includegraphics[width=0.23\textwidth]{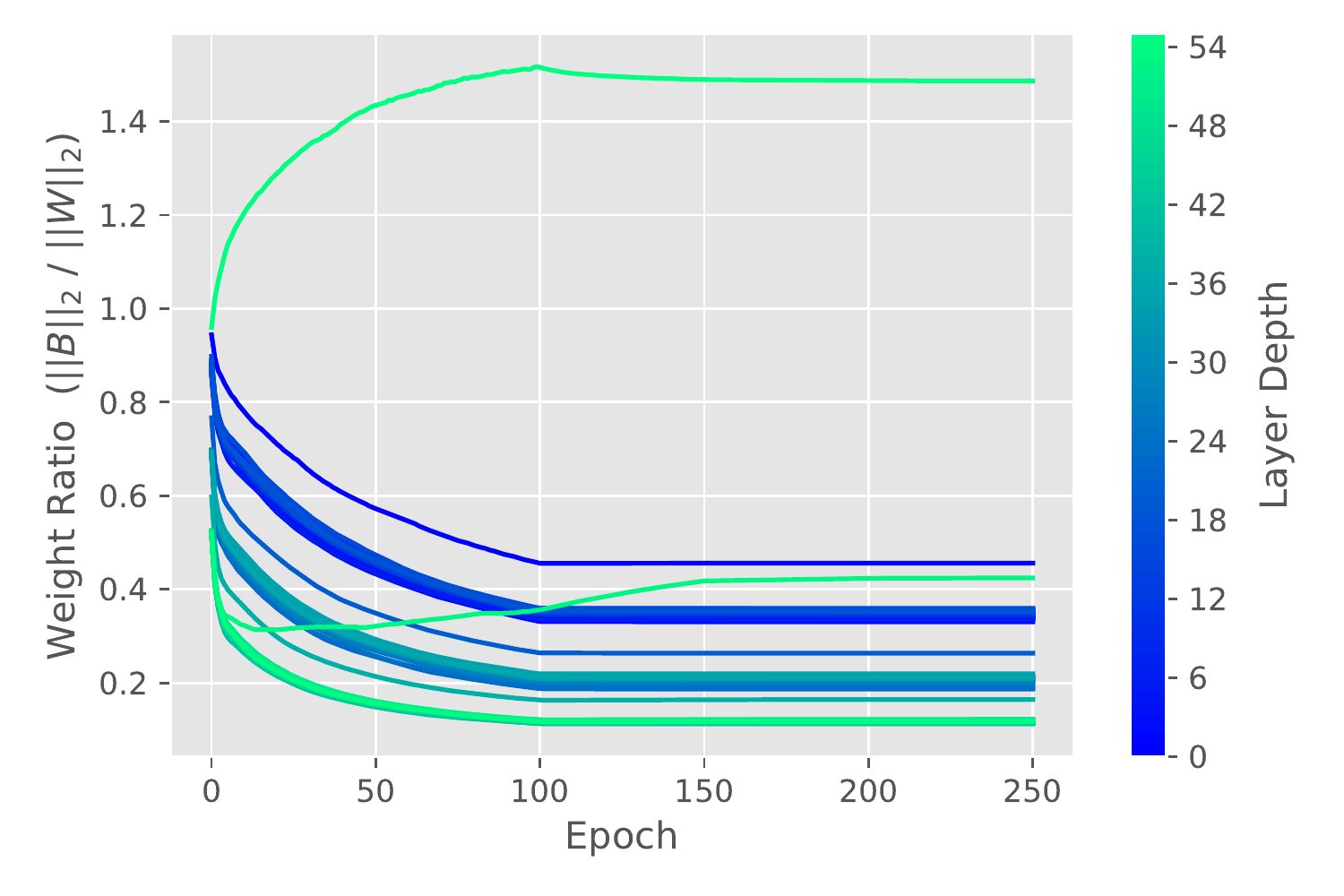}} 
    \caption{Weight ratios for the methods FA and DFA when training a ResNet56 on CIFAR-10 for SGD in (a) and Adam in (b).}
    \label{fig:dfa_ratio_r56}
\end{figure}

The significant improvement brought by Adam is expected since it maintains per-parameter learning rates that are adapted based on the average of the second moments of the gradients inherited from RMSProp. Therefore, it yields better performances under noisy gradients. To confirm this observation, we plot in Fig.~\ref{fig:dfa_ratio_r56} the backward-forward norm weight ratios for the DFA method for both SGD and Adam optimizers. There, it is seen that SGD has driven the norm weight ratios of the first layers of the network very close to 0, which means that the forward weight matrices $\mathbf{W}_i$ were updated to reach much larger values than the ones in the backward weight matrices $\mathbf{B}_i$. This is due to the direct error projection of the error from the last layer to each layer in the DFA method, thereby sidestepping the small norms of the gradients computed using chain rule.

\begin{figure}[h!]
    \centering
    \subfigure[LeNet (SGD)]{
        \includegraphics[width=0.235\textwidth]{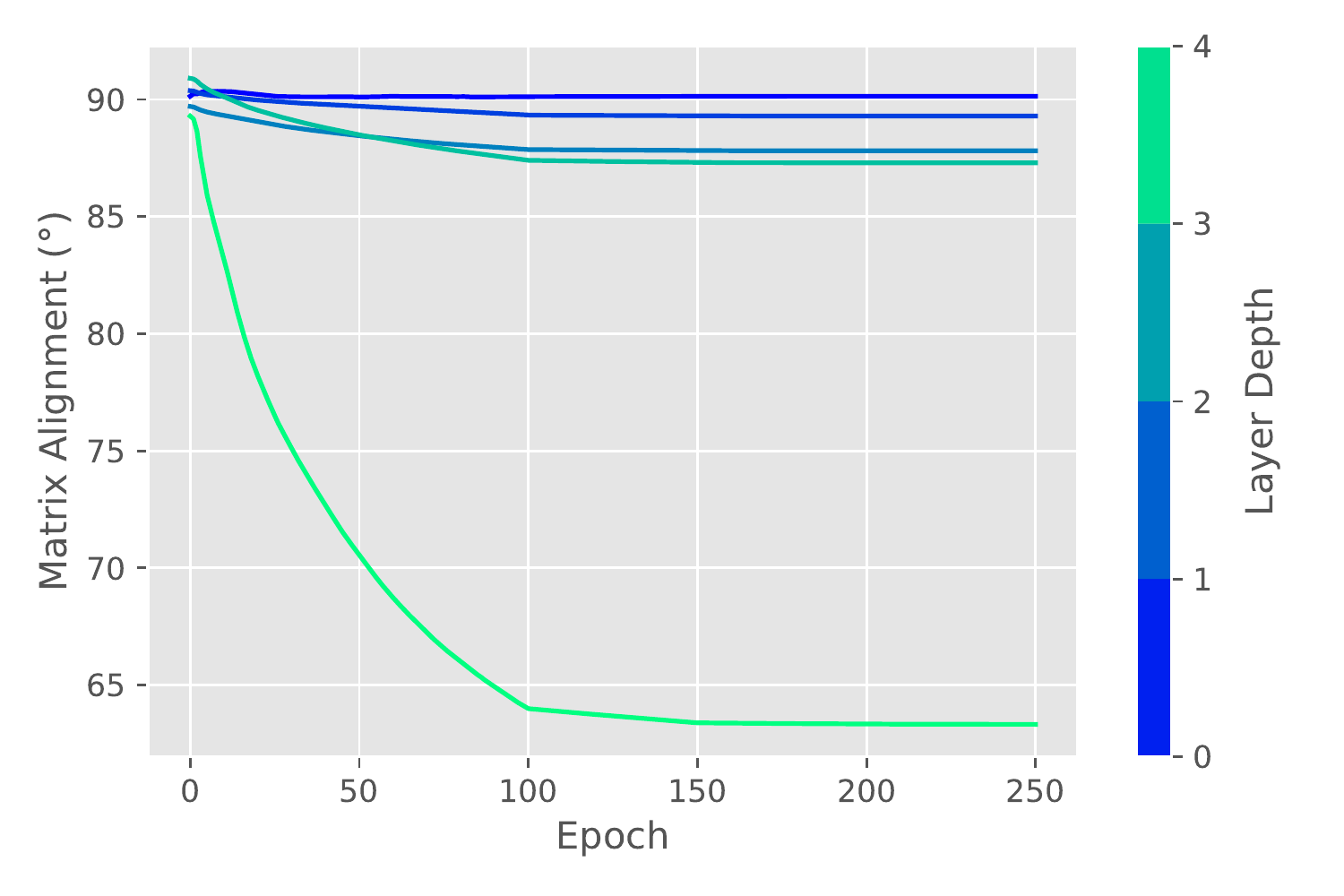}
        \includegraphics[width=0.235\textwidth]{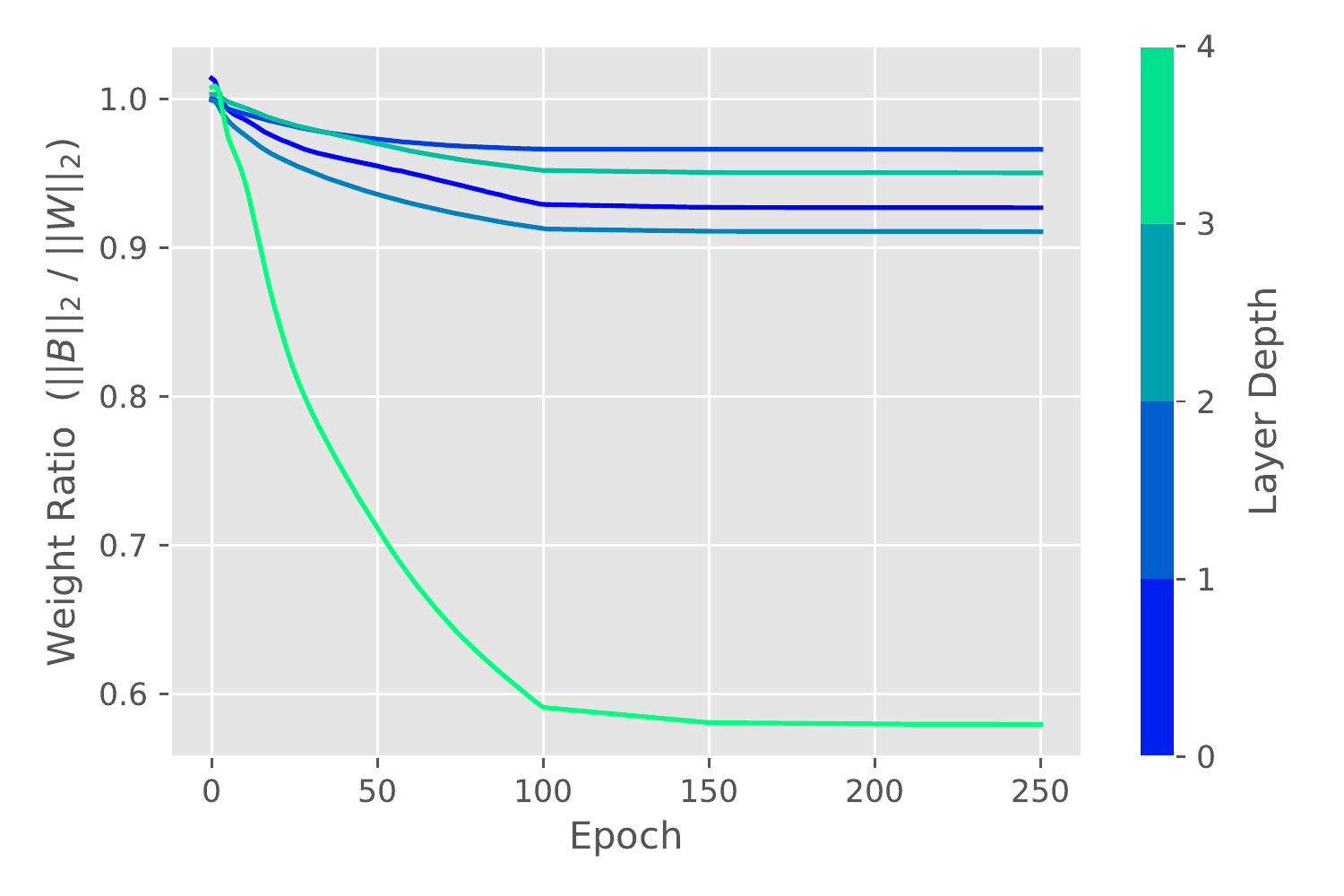}
    }
    \subfigure[ResNet-20 (SGD)]{
        \includegraphics[width=0.235\textwidth]{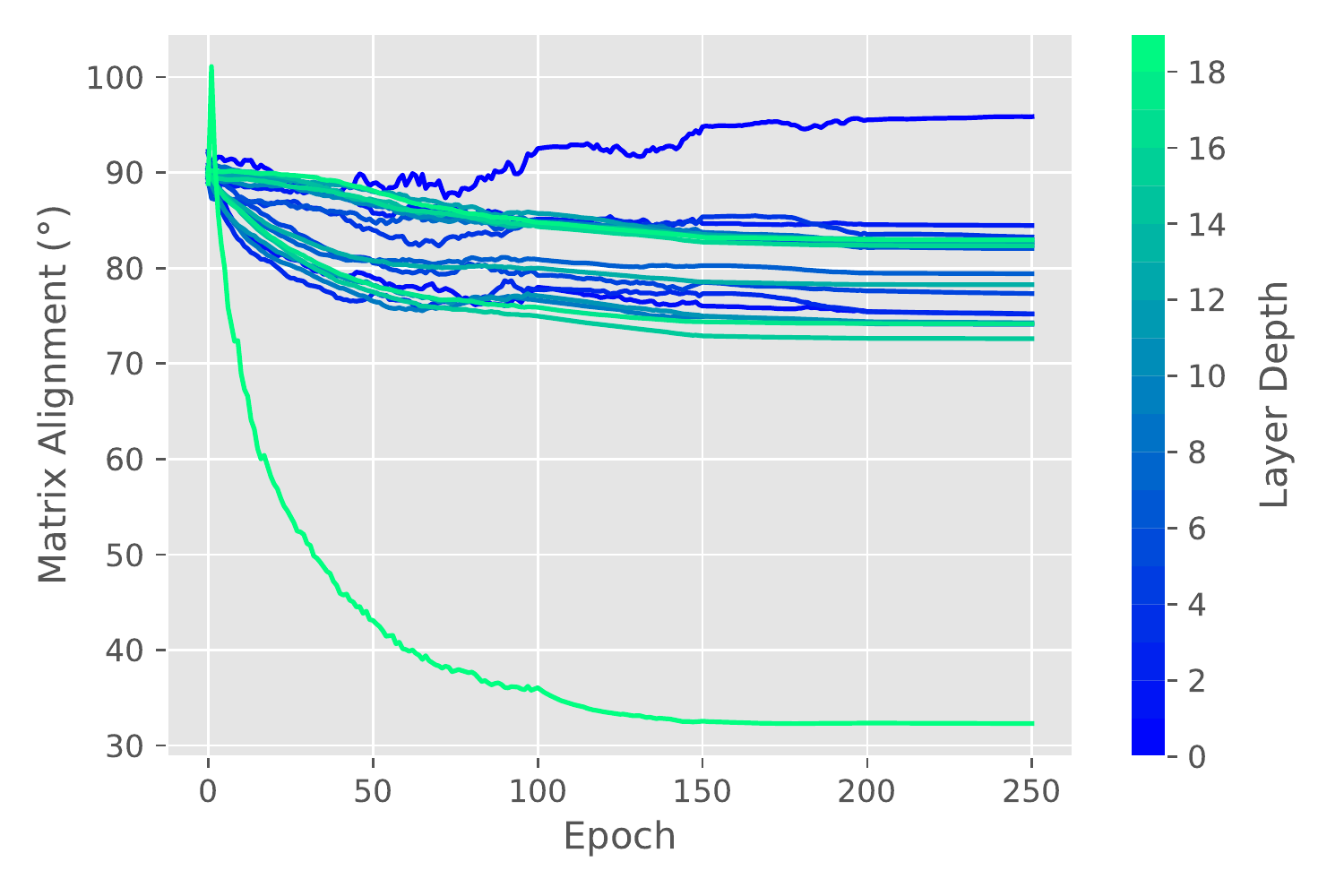}
        \includegraphics[width=0.235\textwidth]{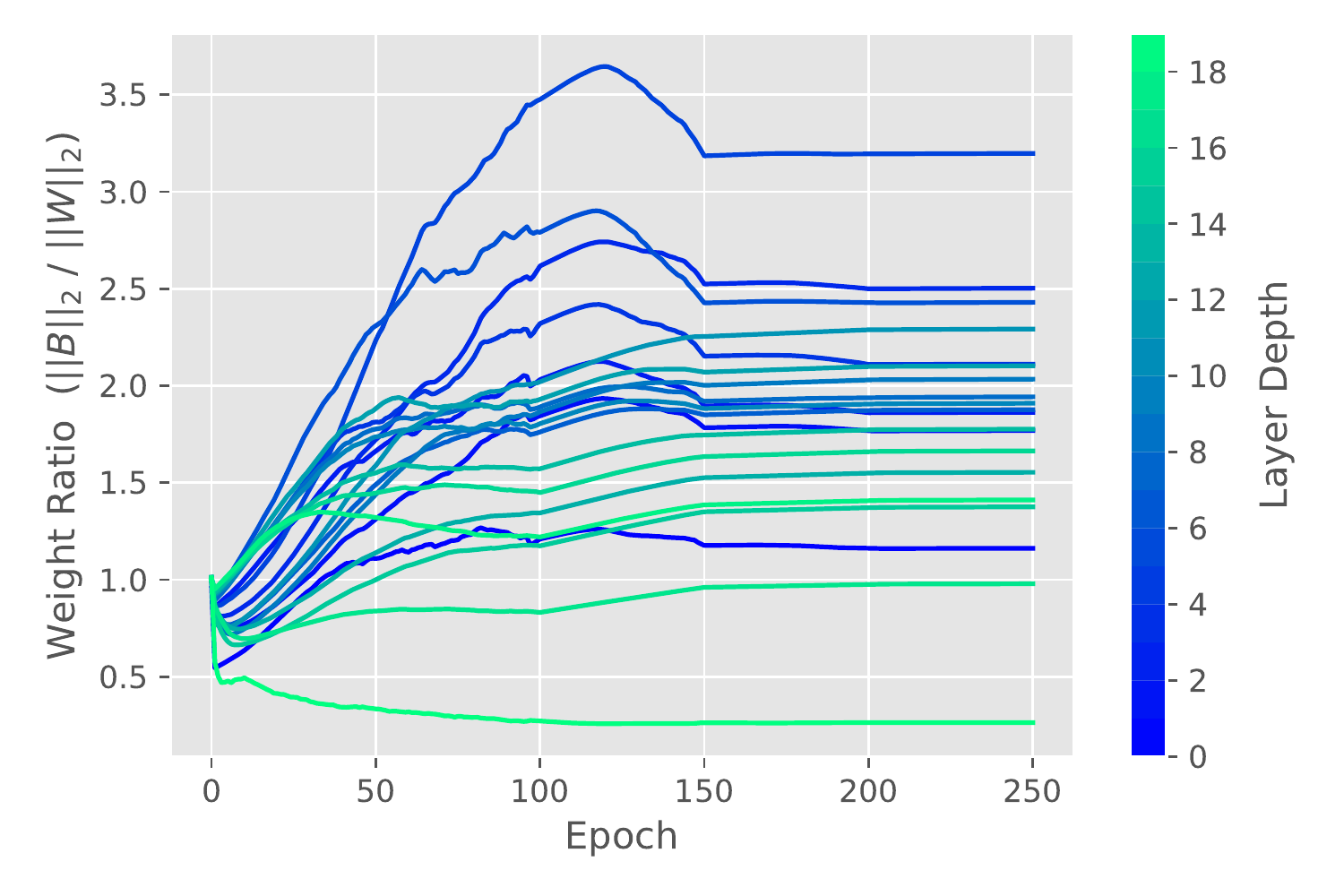}
        }
    \subfigure[LeNet (Adam)]{
        \includegraphics[width=0.235\textwidth]{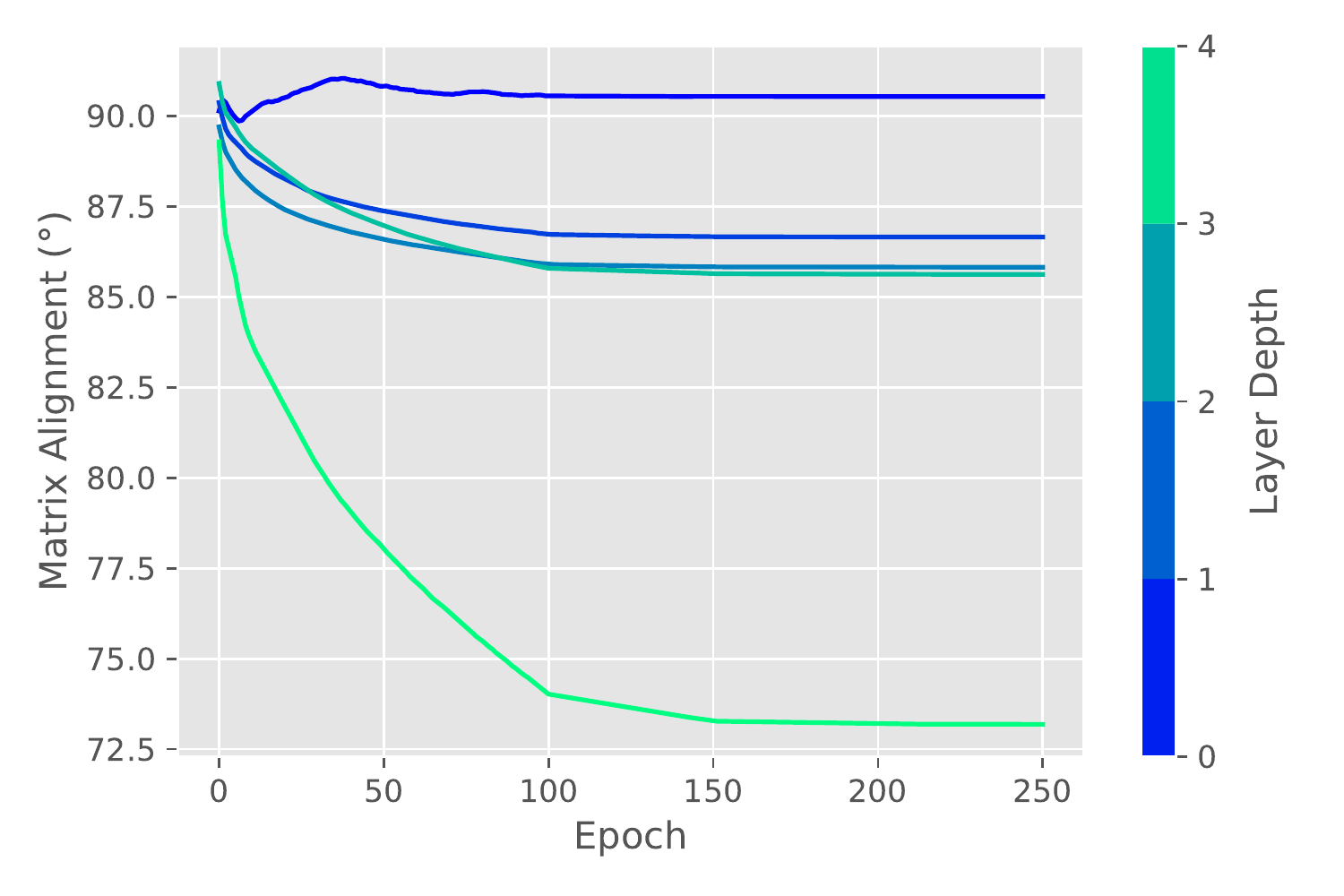}
        \includegraphics[width=0.235\textwidth]{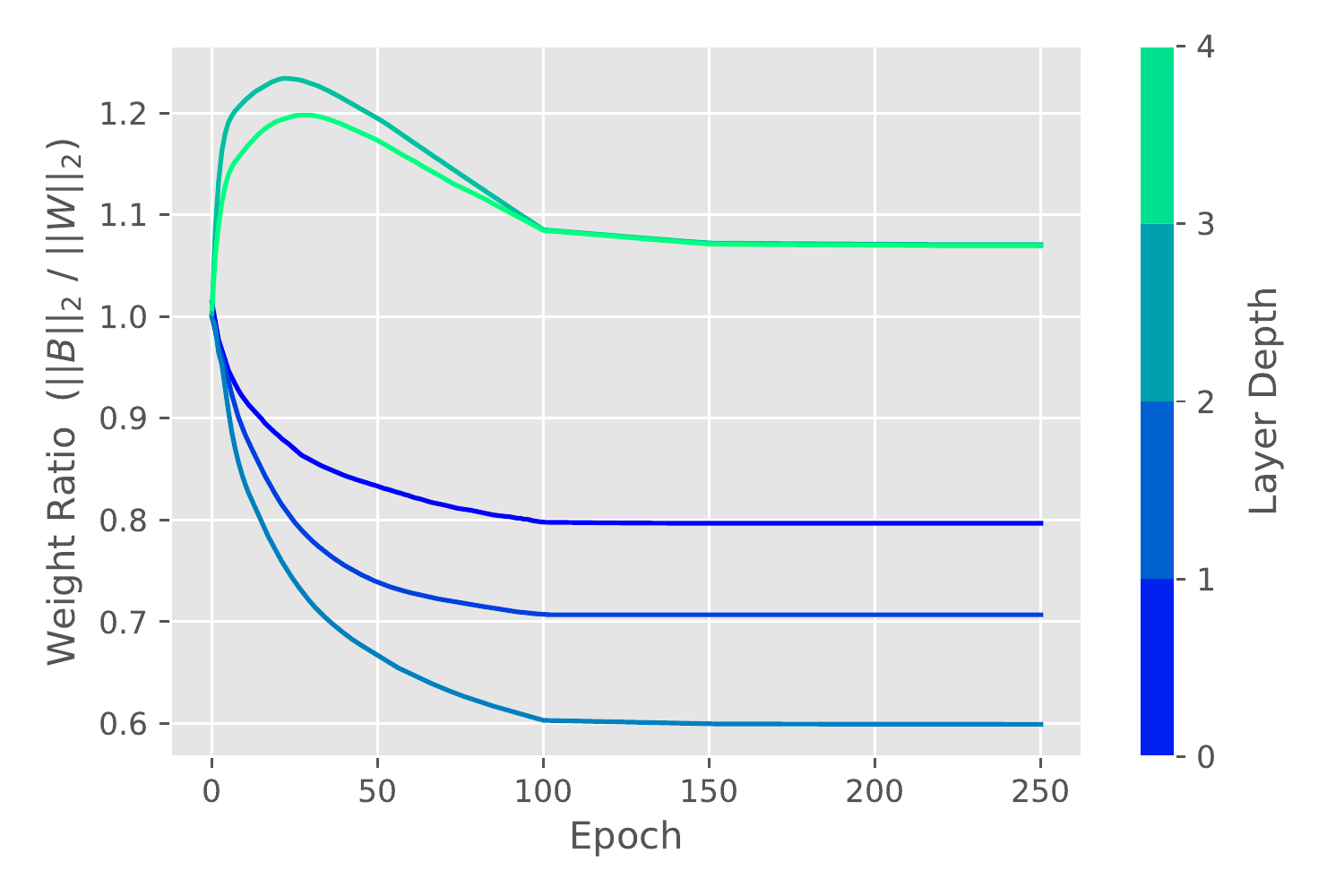}
        } 
    \subfigure[ResNet-20 (Adam)]{
        \includegraphics[width=0.235\textwidth]{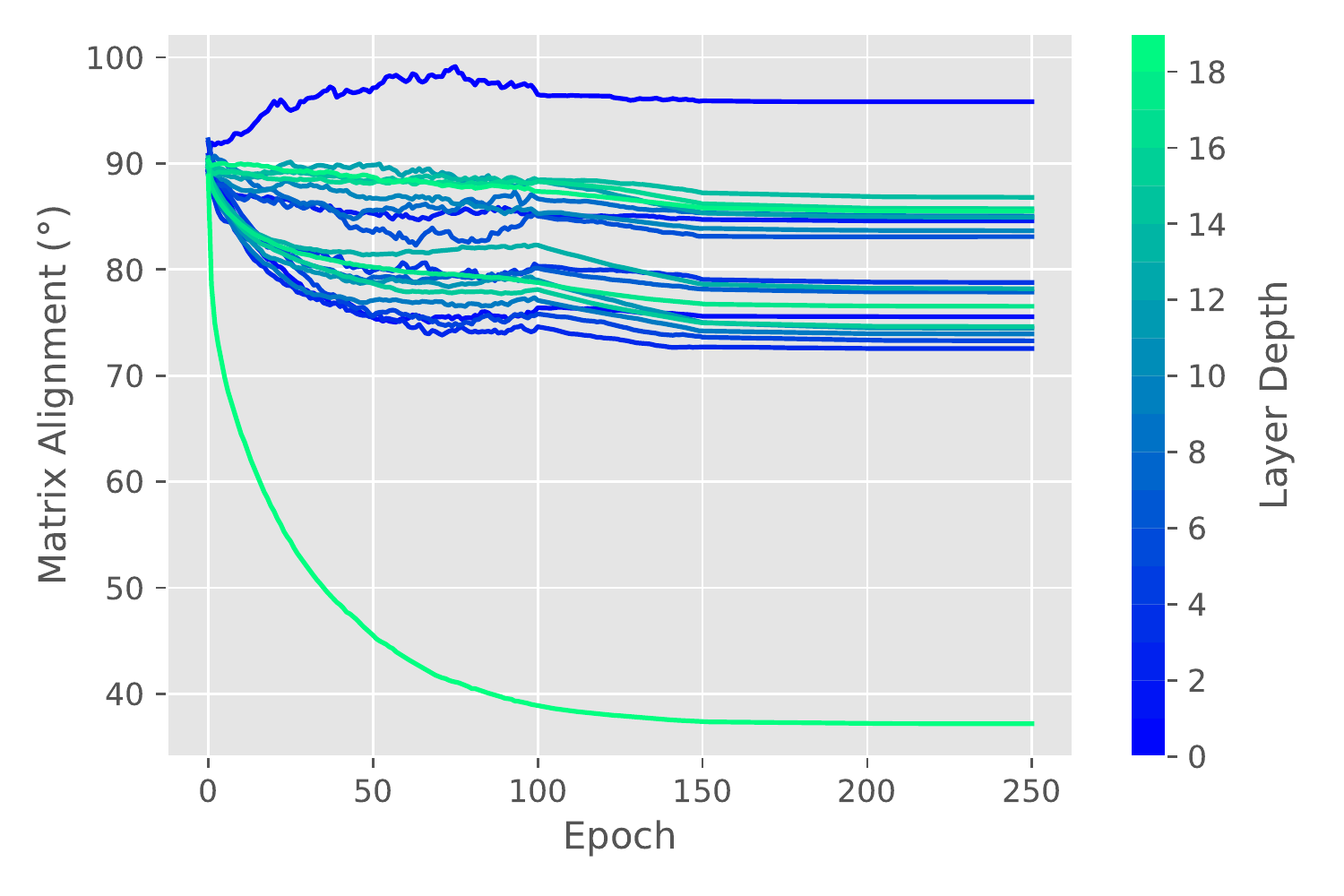}
        \includegraphics[width=0.235\textwidth]{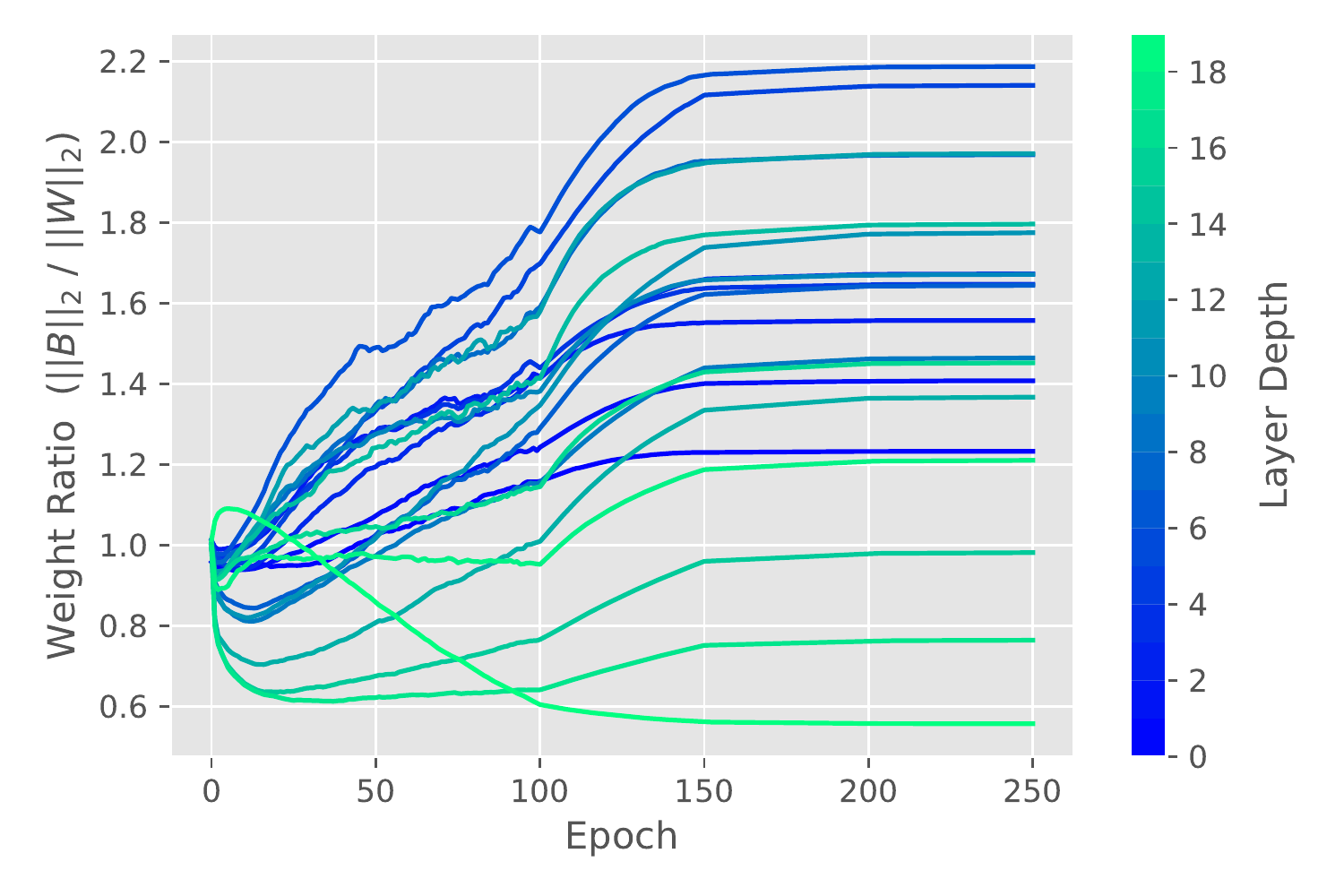}
        }
    \caption{Matrix alignment and weight ratios for FA and different network and optimizer configurations. LeNet with SGD or Adam optimizer in (a) and (c). ResNet-20 with SGD or Adam optimizer in (b) and (d)}
    \label{fig:layer_alignment_all}
\end{figure}

The same observation does not hold for the FA method, where the weight norm ratio of the first layers does not vanish to 0 as shown in Fig.~\ref{fig:layer_alignment_all}. We also see that Adam achieves smaller backward-forward weight alignment angles compared to SGD for both LeNet and ResNet-20.

\begin{figure}[h!]
    \centering
    \subfigure[Kaiming (SGD)]{
        \includegraphics[width=0.235\textwidth]{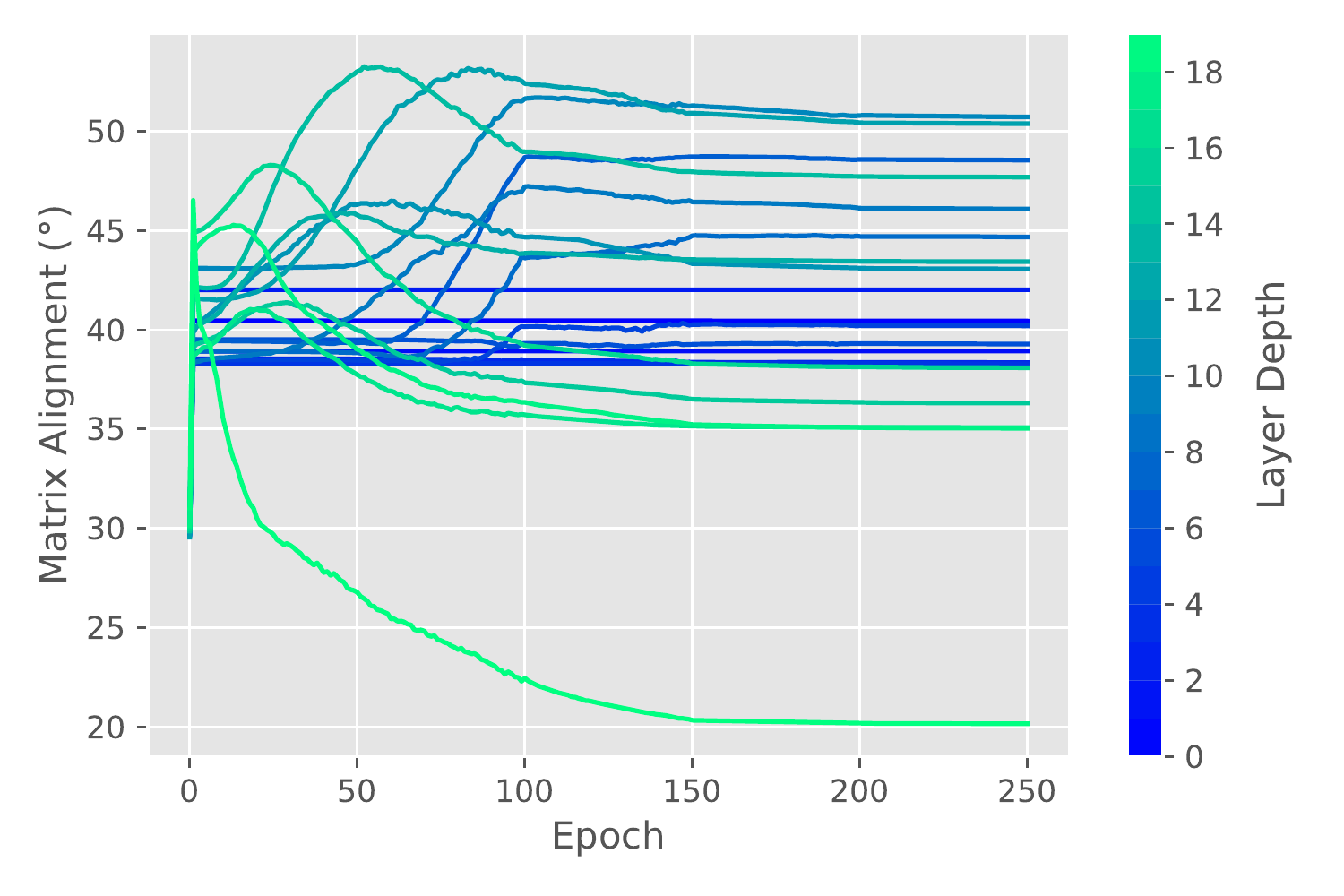}
        \includegraphics[width=0.235\textwidth]{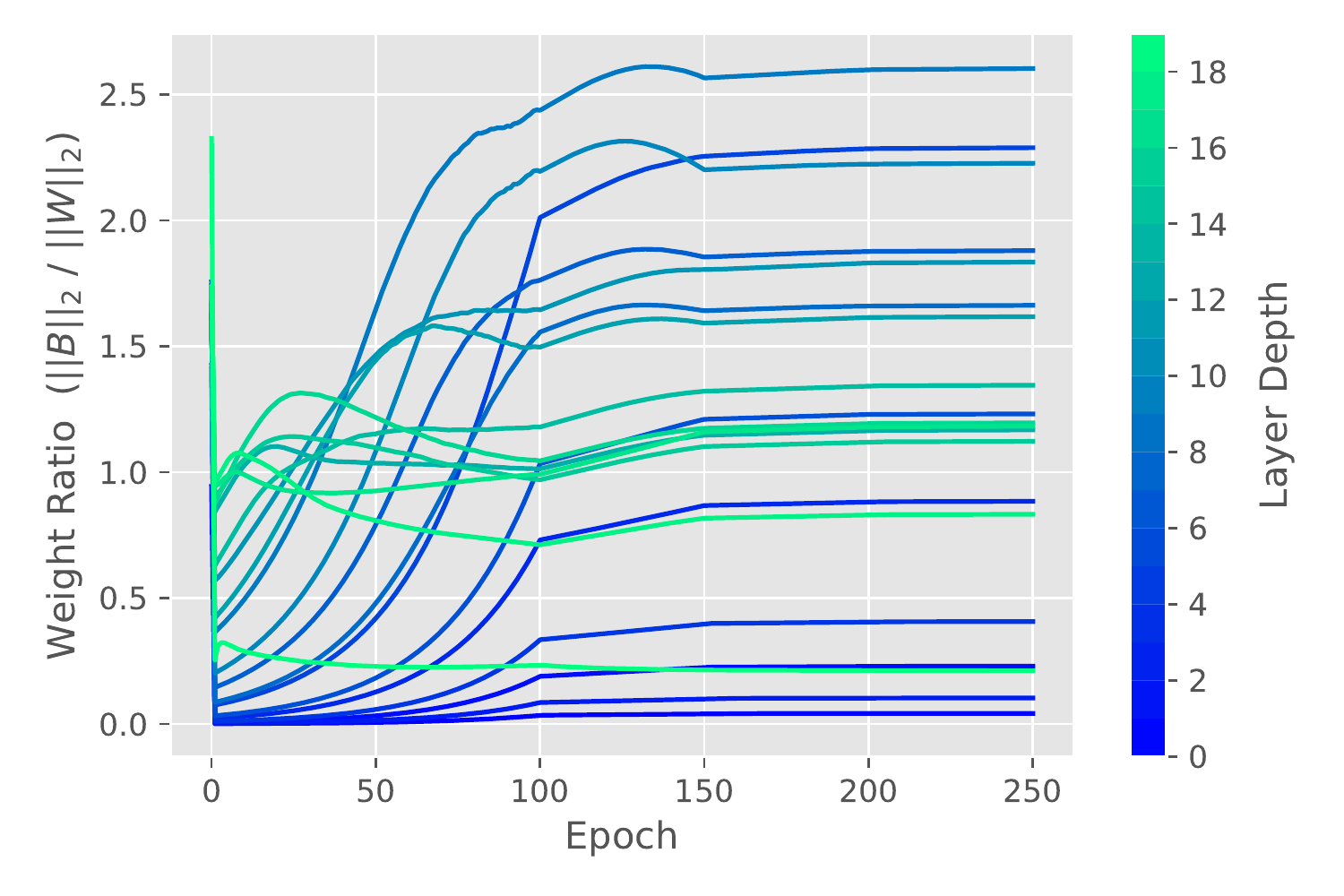}
    }
    \subfigure[$\mathcal{N}(0, 1)$ (SGD)]{
        \includegraphics[width=0.235\textwidth]{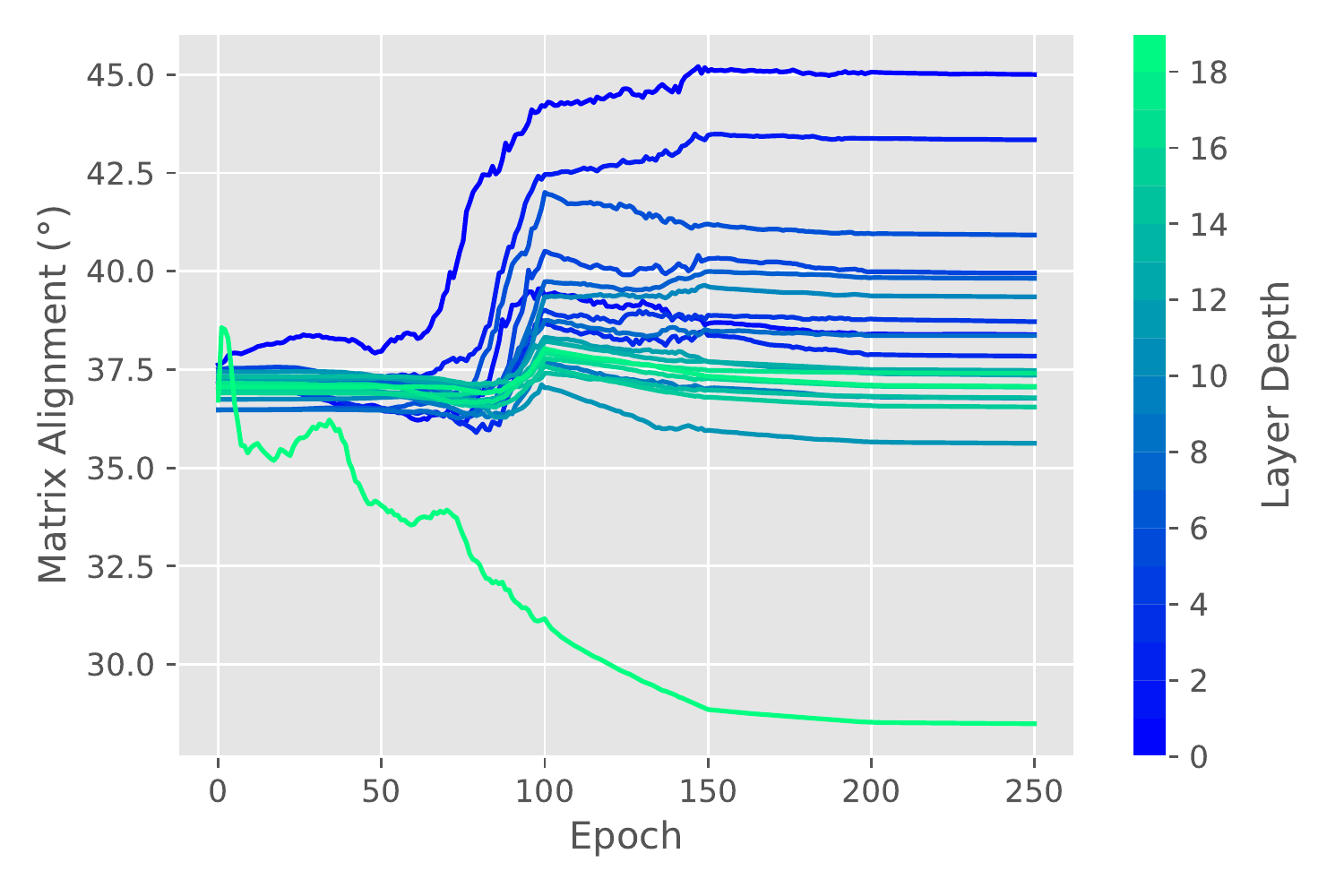}
        \includegraphics[width=0.235\textwidth]{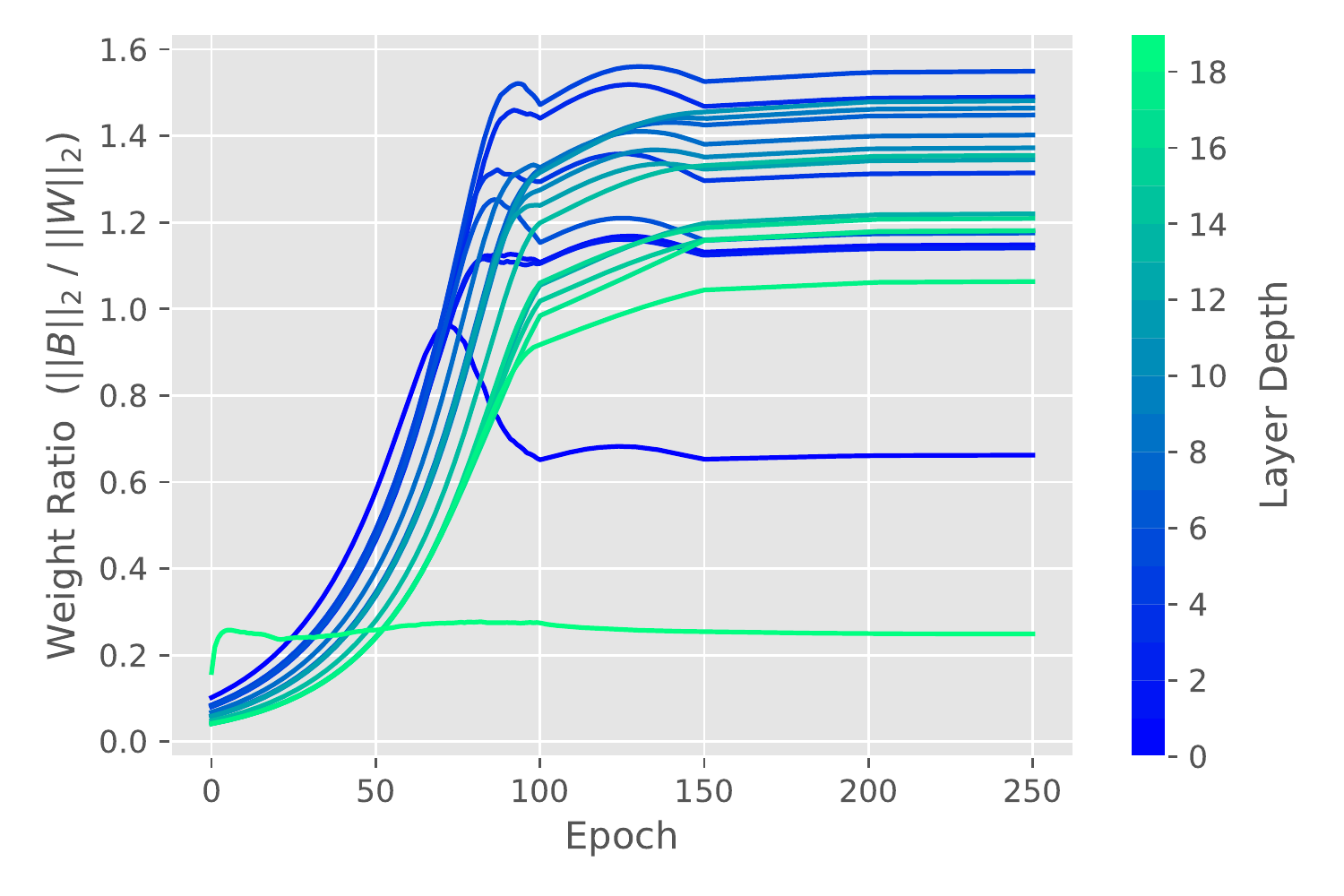}
    }
    \subfigure[Kaiming (Adam)]{
        \includegraphics[width=0.235\textwidth]{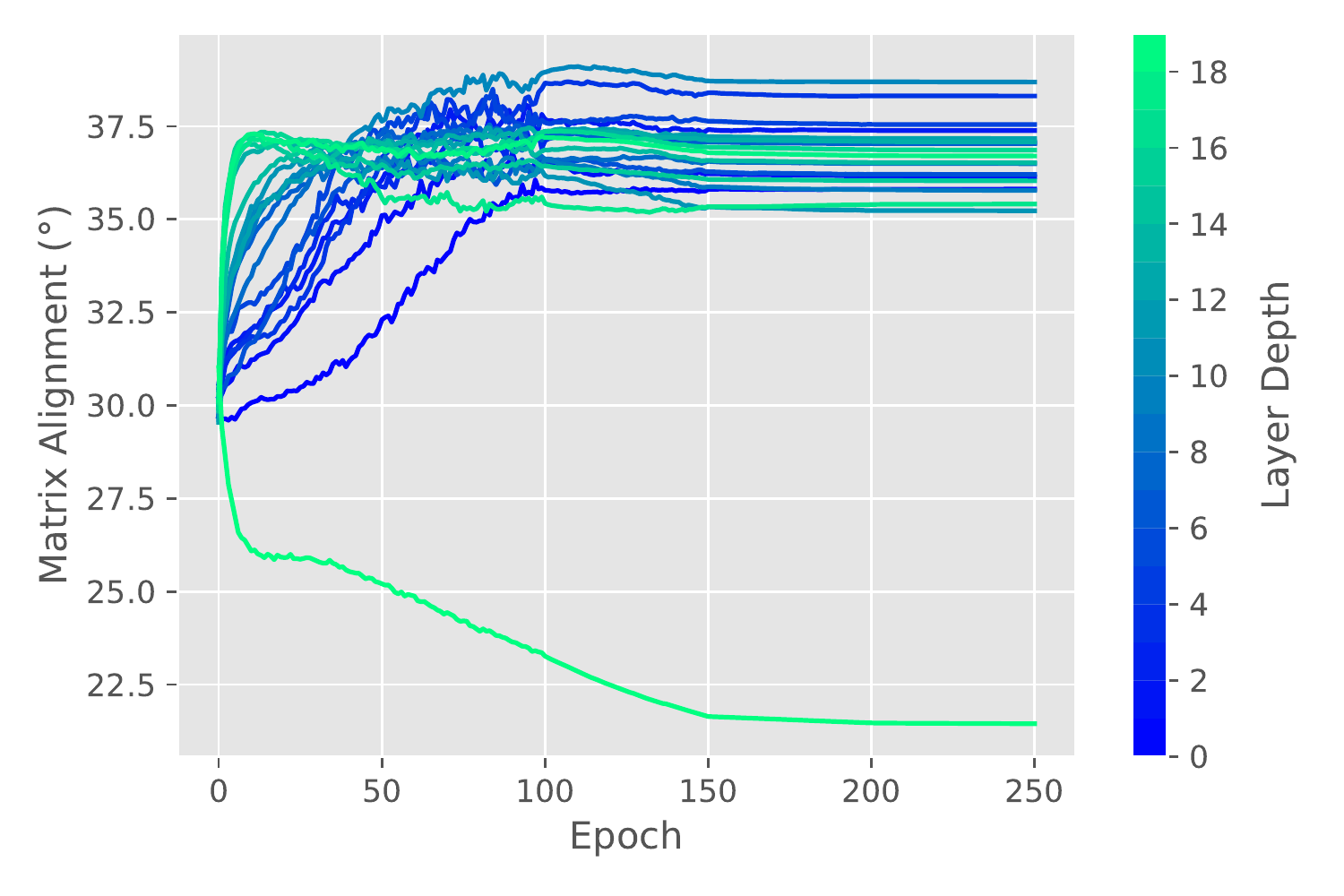}
        \includegraphics[width=0.235\textwidth]{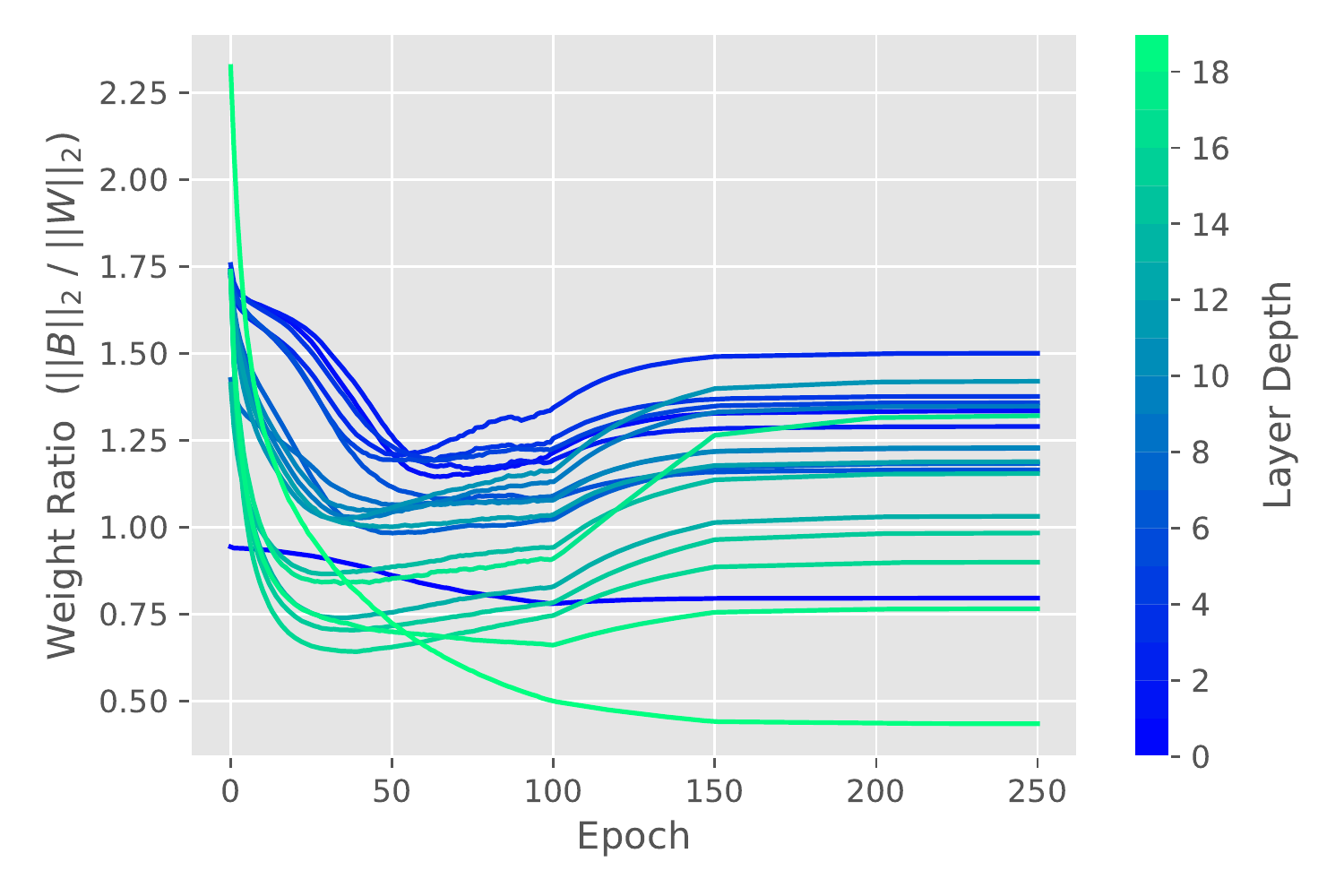}
    }
    \subfigure[$\mathcal{N}(0, 1)$ (Adam)]{
        \includegraphics[width=0.235\textwidth]{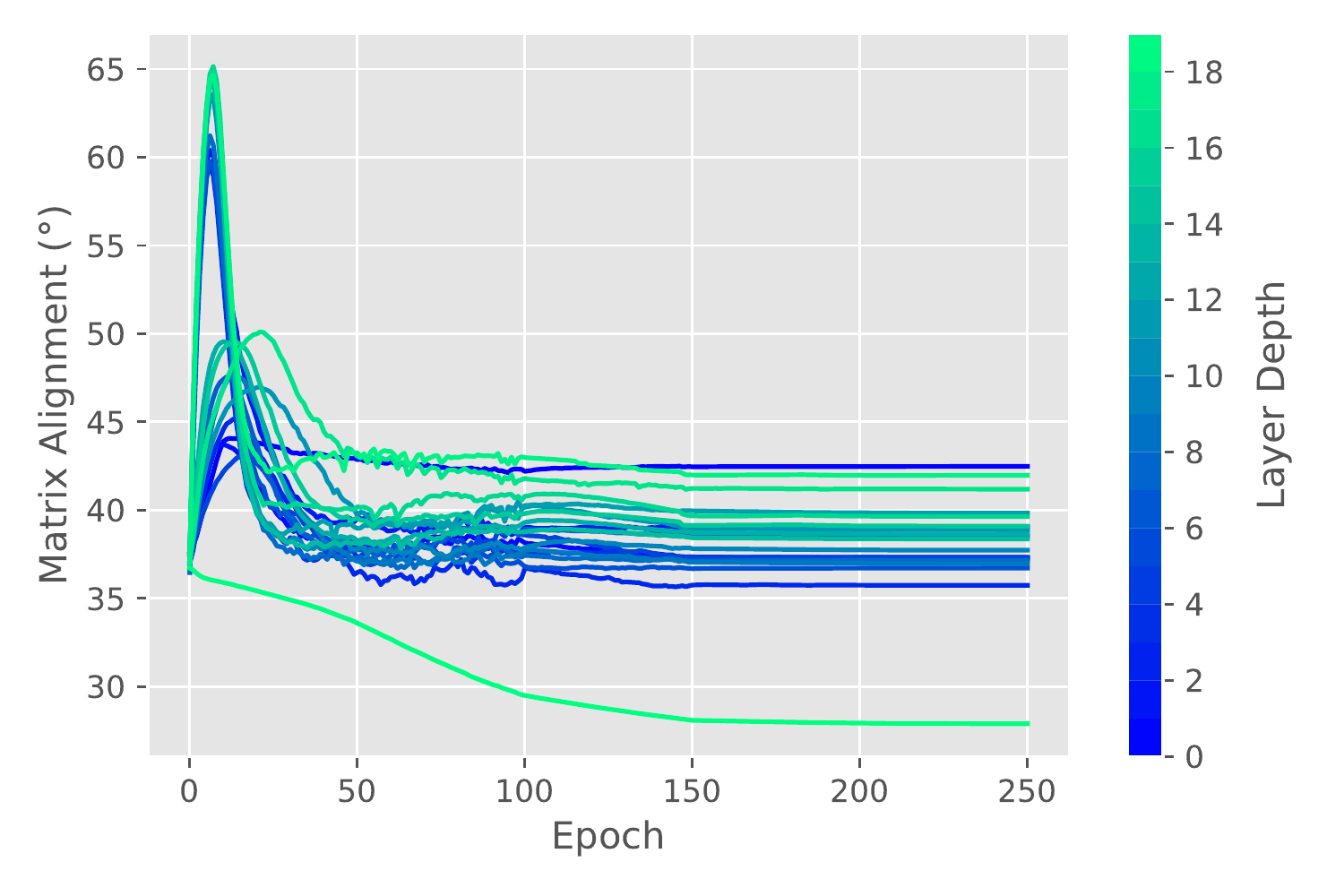}
        \includegraphics[width=0.235\textwidth]{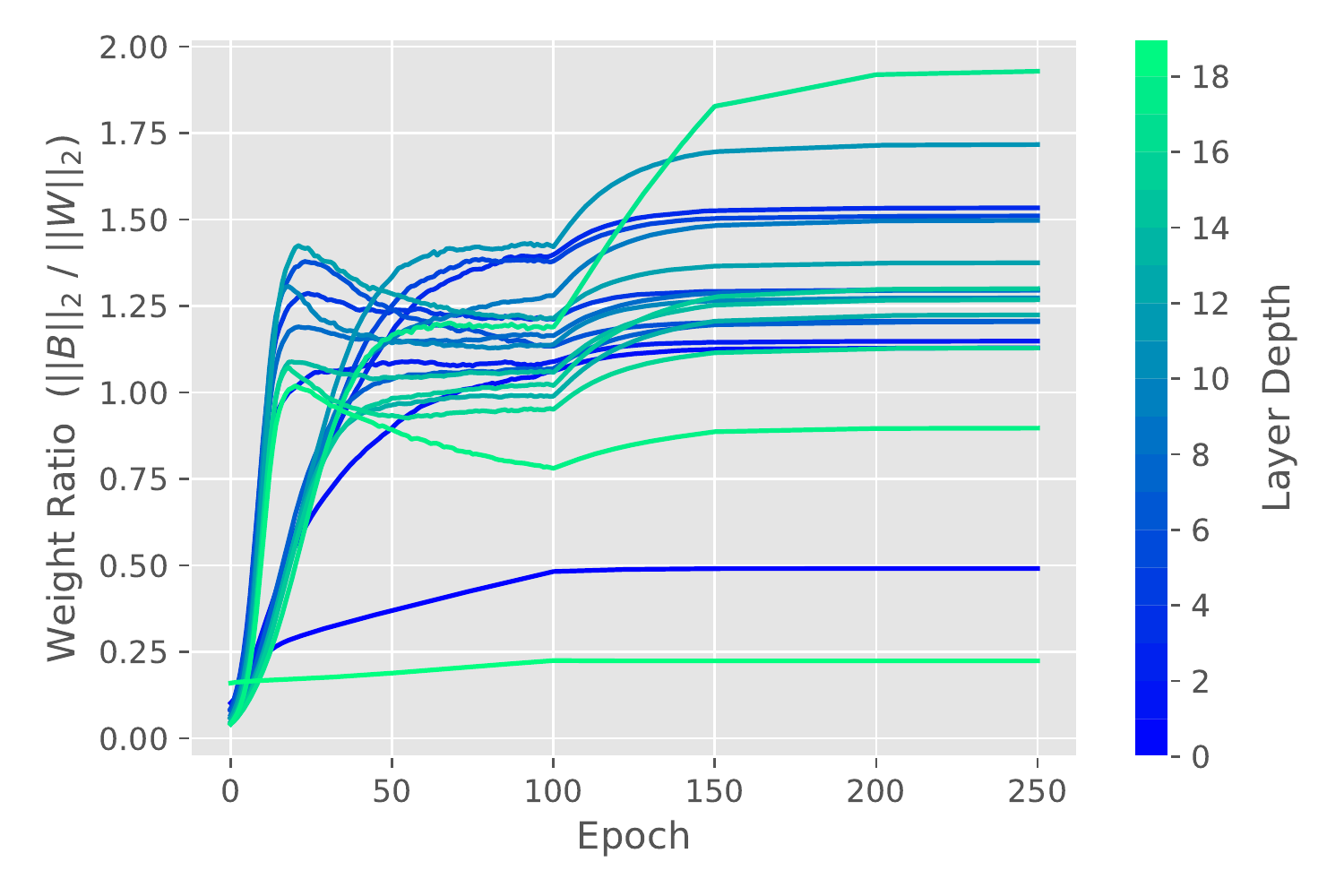}
    }
    \caption{Matrix alignment and weight ratios for uSF for a ResNet-20 network trained with SGD and Adam on CIFAR10. The forward weight matrices are initialized with Kaiming in (a) and (c) and with $\mathcal{N}(0,1)$ in (b) and (d).}
    \label{fig:usf_diff_inits_r20_angles_weights}
\end{figure}

Furthermore, we investigate the effect of the difference between the distributions used to initialize the forward and backward weights, $\mathbf{W}_i$ and $\mathbf{B}_i$, respectively. To this end, we initialized the backward weights with the Xavier initialization. Then, we run two separate experiments in which we initialized the forward weights with a Kaiming Uniform \footnote{\textit{PyTorch} default Kaiming init. is defined as \texttt{nn.init.kaiming\_uniform\_(weight, a=math.sqrt(5))}} (\textit{PyTorch} default layer initialization), and then with a Normal distribution $\mathcal{N}(0, 1)$. The results are shown in Fig.~\ref{fig:usf_diff_inits_r20_angles_weights}.

We noticed how Adam was able to get a better alignment faster than SGD, and outperforming it in accuracy when there was a mismatch between the initialization methods. Furthermore, we observed that the mismatch of initialization compounds with network depth, making it more difficult to reach convergence as more layers are added. 

Finally, we benchmark all the methods training a ResNet-18 network on ImageNet. Our results in Fig.~\ref{fig:imagenet_error} and Table~\ref{fig:imagenet_error} showing the top-1 classification error rate in the validation set, comply with the values previously reported in~\cite{xiao2018biologically}. The training details are described in the Appendix Section~\ref{sec:experiment_details}.

\begin{table}[h!]
	\begin{minipage}{0.3\linewidth}
		\centering
		\scriptsize
		\begin{tabular}{@{}cc@{}}
\cmidrule(l){2-2}
 & Top -1 Error Rate (\%) \\ \midrule
BP     & 30.39                  \\ \midrule
FA     & 85.25                  \\ \midrule
DFA    & 82.45                  \\ \midrule
uSF    & 34.97                  \\ \midrule
brSF   & 37.21                  \\ \midrule
frSF   & 36.5                   \\ \bottomrule
\end{tabular}
\vspace{0.2cm}
\caption{Top-1 ImageNet validation error for all the alignment methods.}
\label{tab:imagenet_error}
	\end{minipage}\hfill
	\begin{minipage}{0.55\linewidth}
		\centering
		\vspace{0.3cm}
		\includegraphics[width=0.7\textwidth]{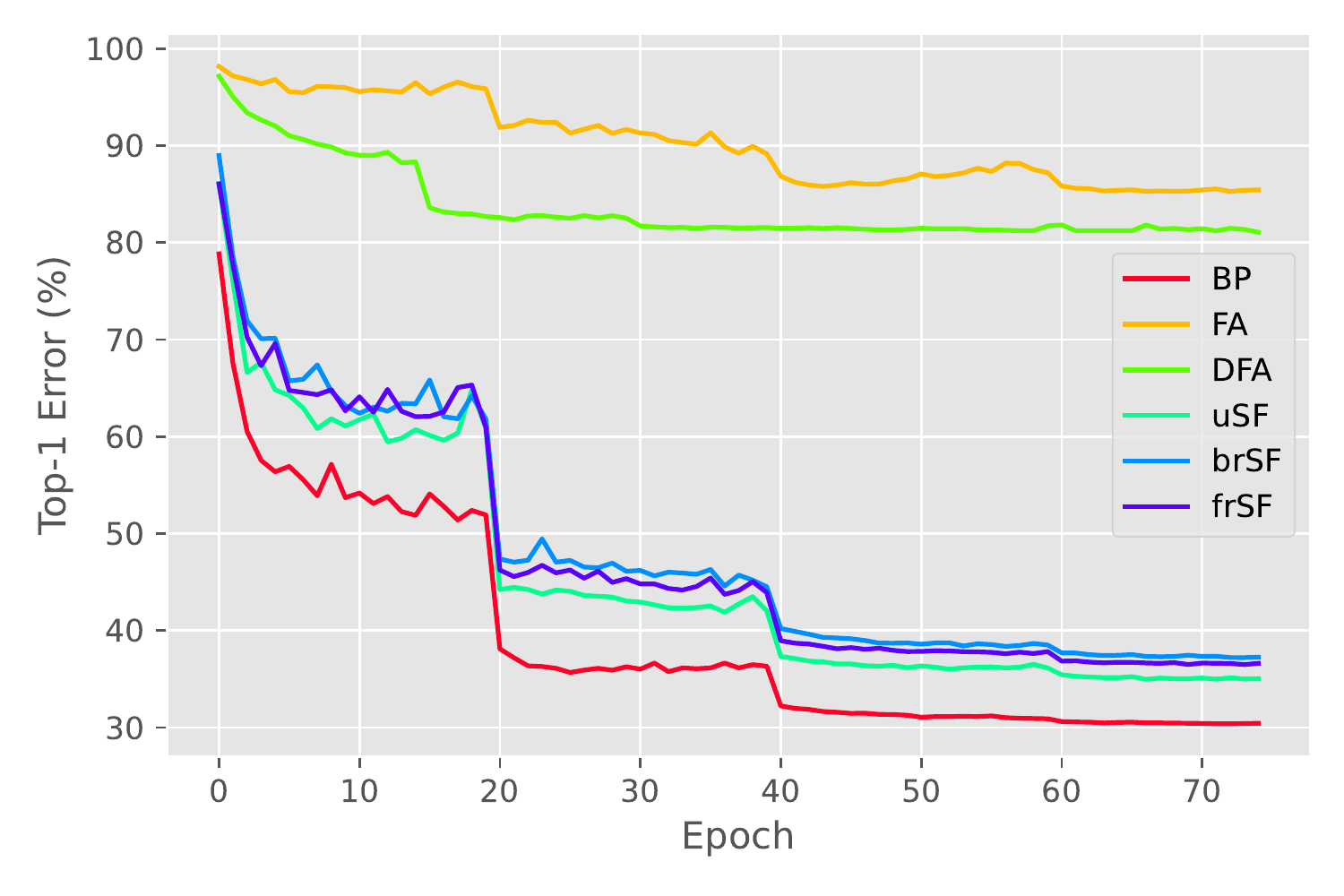}
     \caption{Top-1 ImageNet validation error (\%) for a ResNet-18 network trained with all the feedback alignment methods.}
    \label{fig:imagenet_error}
	\end{minipage}
\end{table}

\subsection{Adversarial Robustness Study}
In this section, we benchmark the robustness of our ResNet-20 models trained on CIFAR-10 against the white-box and black-box attacks described in Appendix Section~\ref{appendix:adversarial-attacks} using the open-source \textit{Torchattacks} package~\cite{kim2020torchattacks}. The evaluation of the white-box attacks was performed on the full CIFAR-10 test set, while the black-box attacks were evaluated on 500 test images only, due to their significant model query complexities.

\begin{figure}[h!]
    \centering
    \scriptsize{\includegraphics[scale=0.8]{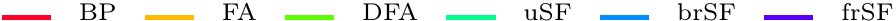}}\\
    \subfigure[{FGSM} (SGD)]{\includegraphics[width=0.22\textwidth]{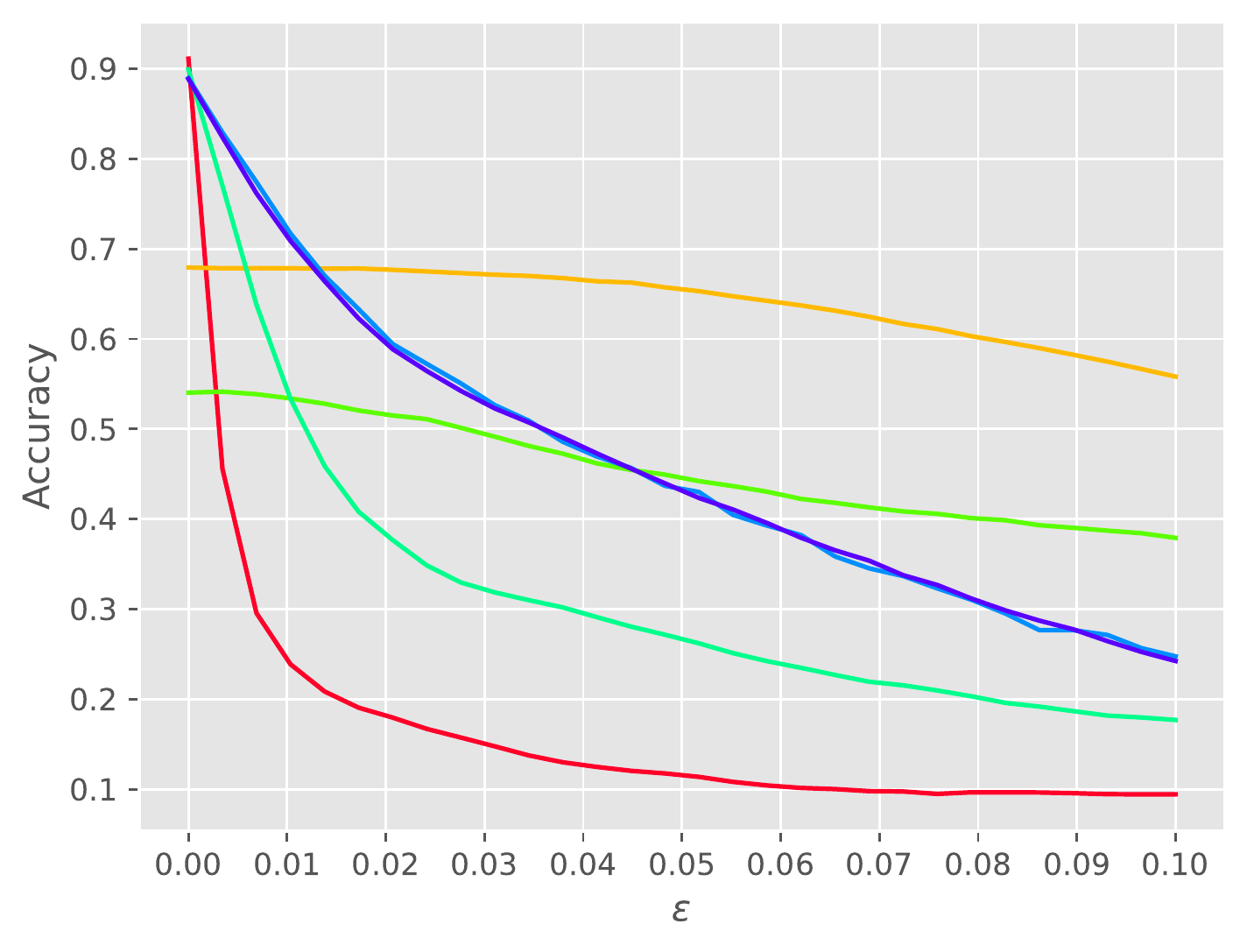}\label{fig:fgsm-sgd}}
    \subfigure[{PGD (SGD)}]{\includegraphics[width=0.22\textwidth]{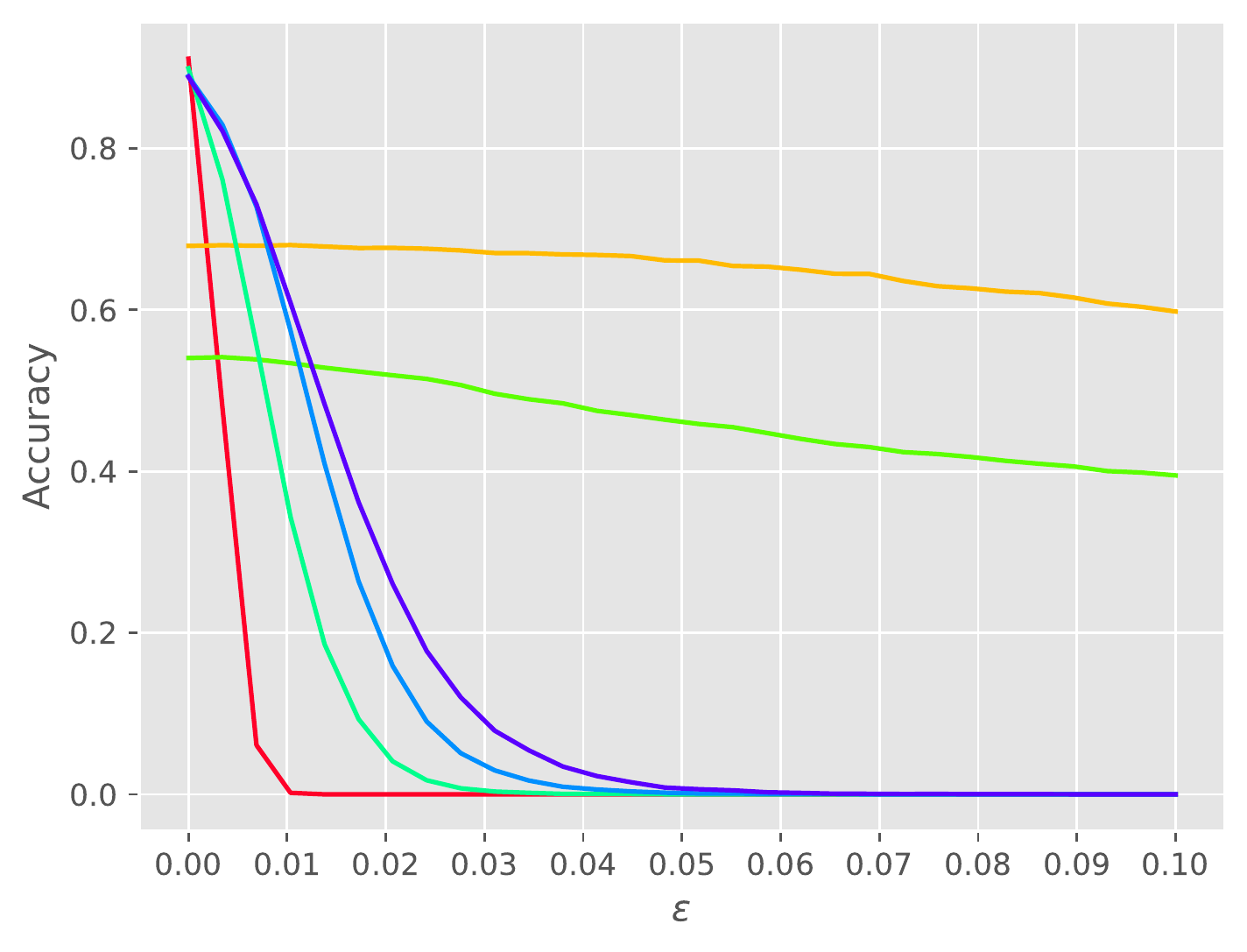}\label{fig:pgd-sgd}}
    \subfigure[{APGD (SGD)}]{\includegraphics[width=0.22\textwidth]{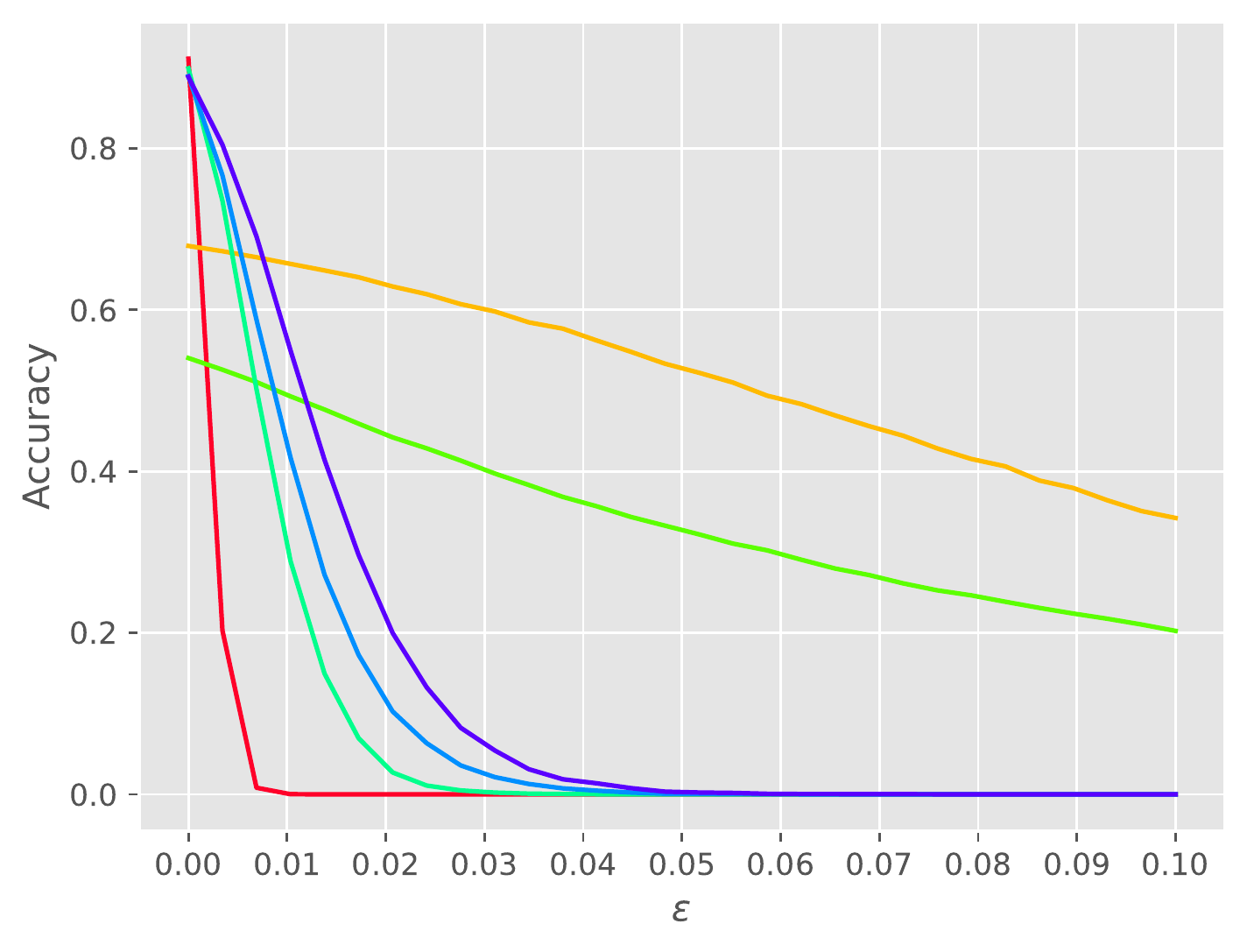}\label{fig:apgd-sgd}}
    \subfigure[TPGD (SGD)]{\includegraphics[width=0.22\textwidth]{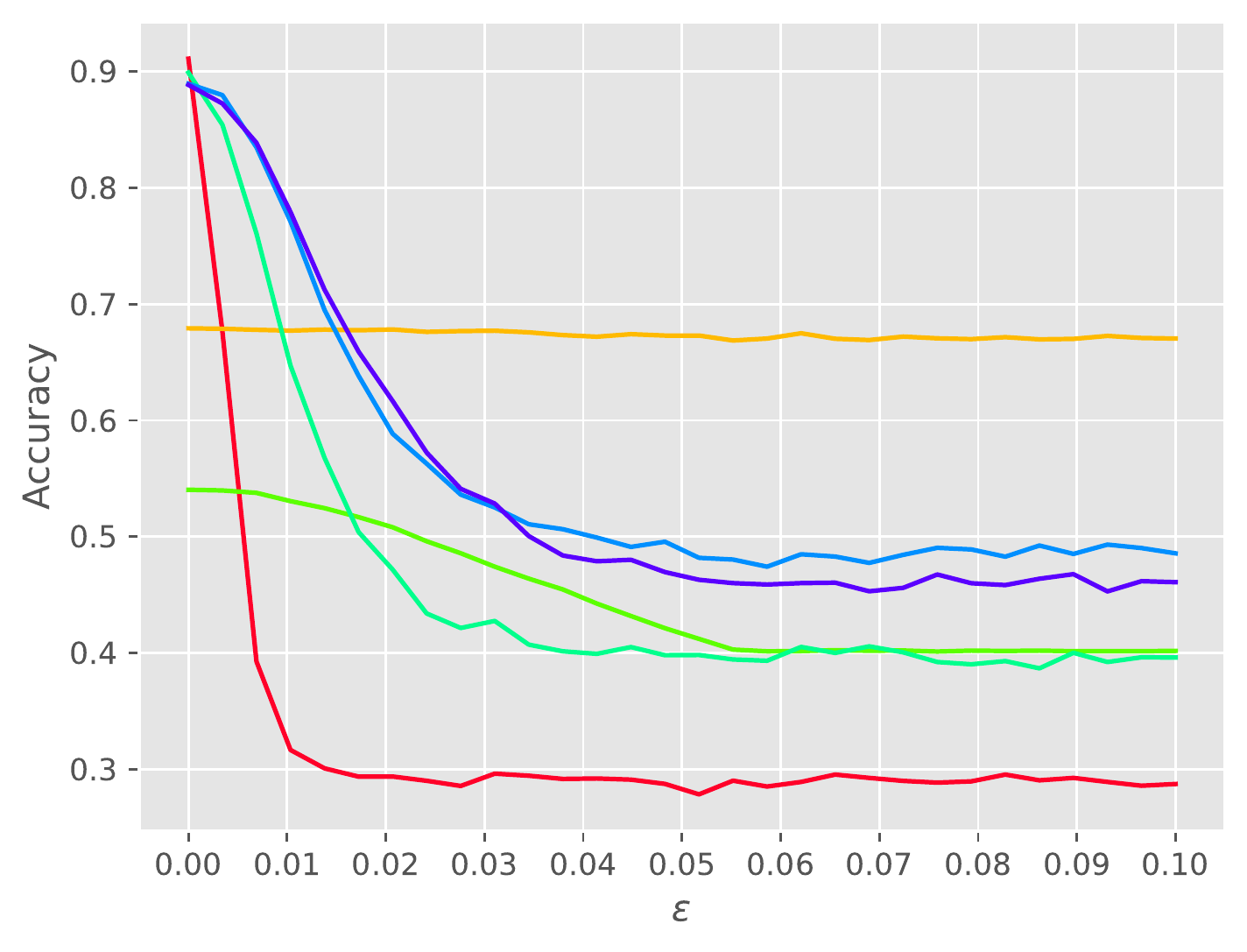}\label{fig:tpgd-sgd}}\\
    \subfigure[FGSM (Adam)]{\includegraphics[width=0.22\textwidth]{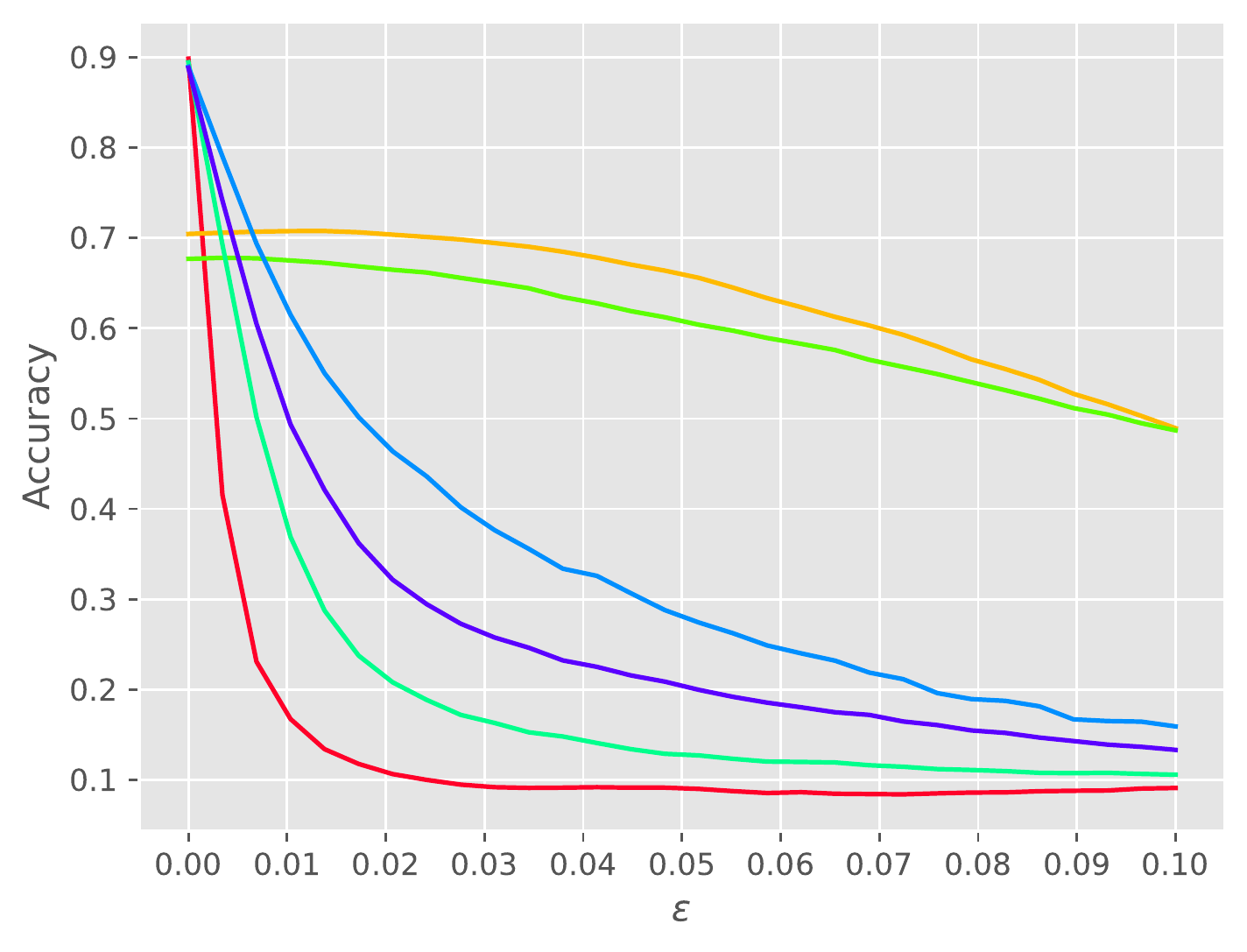}\label{fig:fgsm-adam}} 
    \subfigure[PGD (Adam)]{\includegraphics[width=0.22\textwidth]{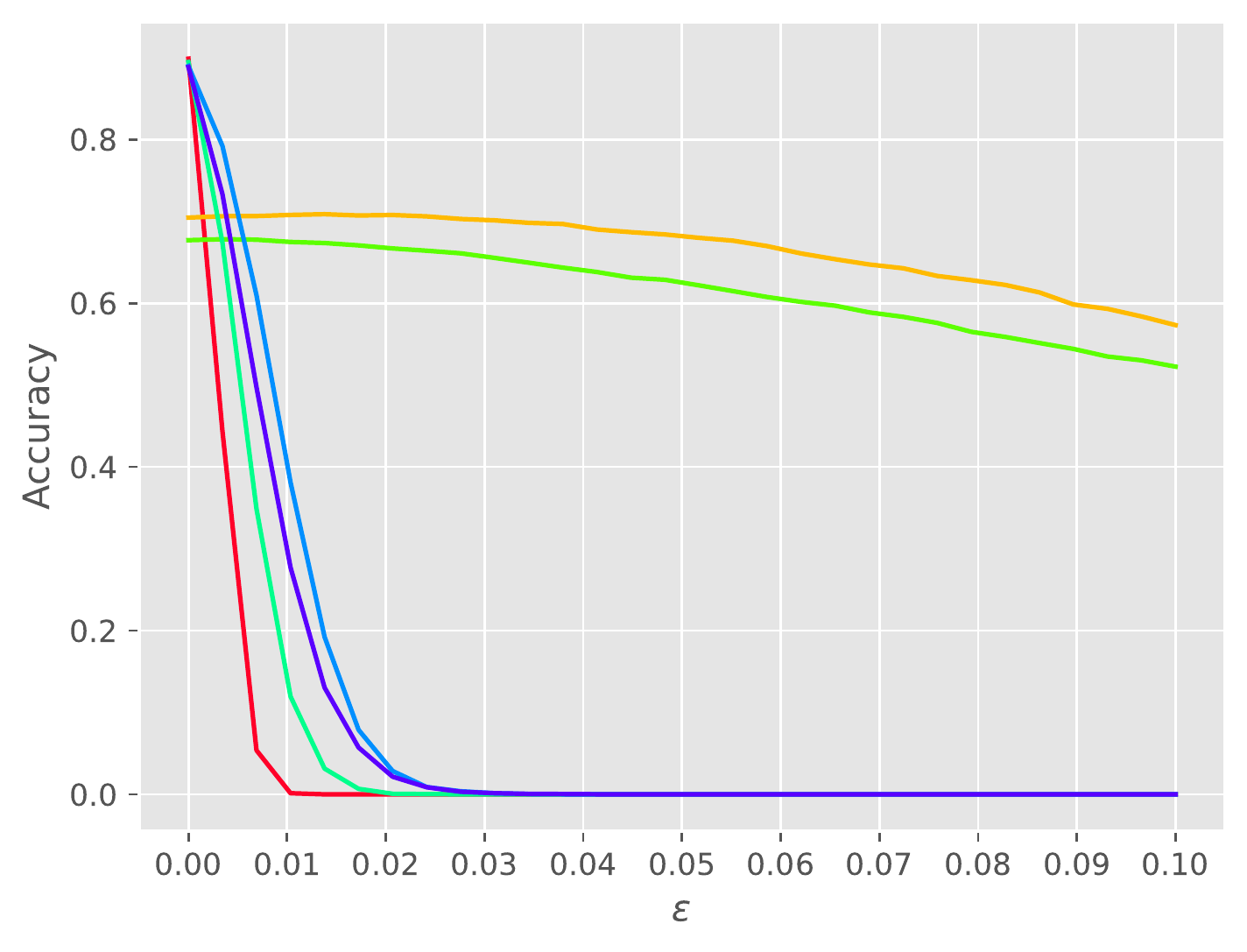}\label{fig:apgd-adam}}
    \subfigure[APGD (Adam)]{\includegraphics[width=0.22\textwidth]{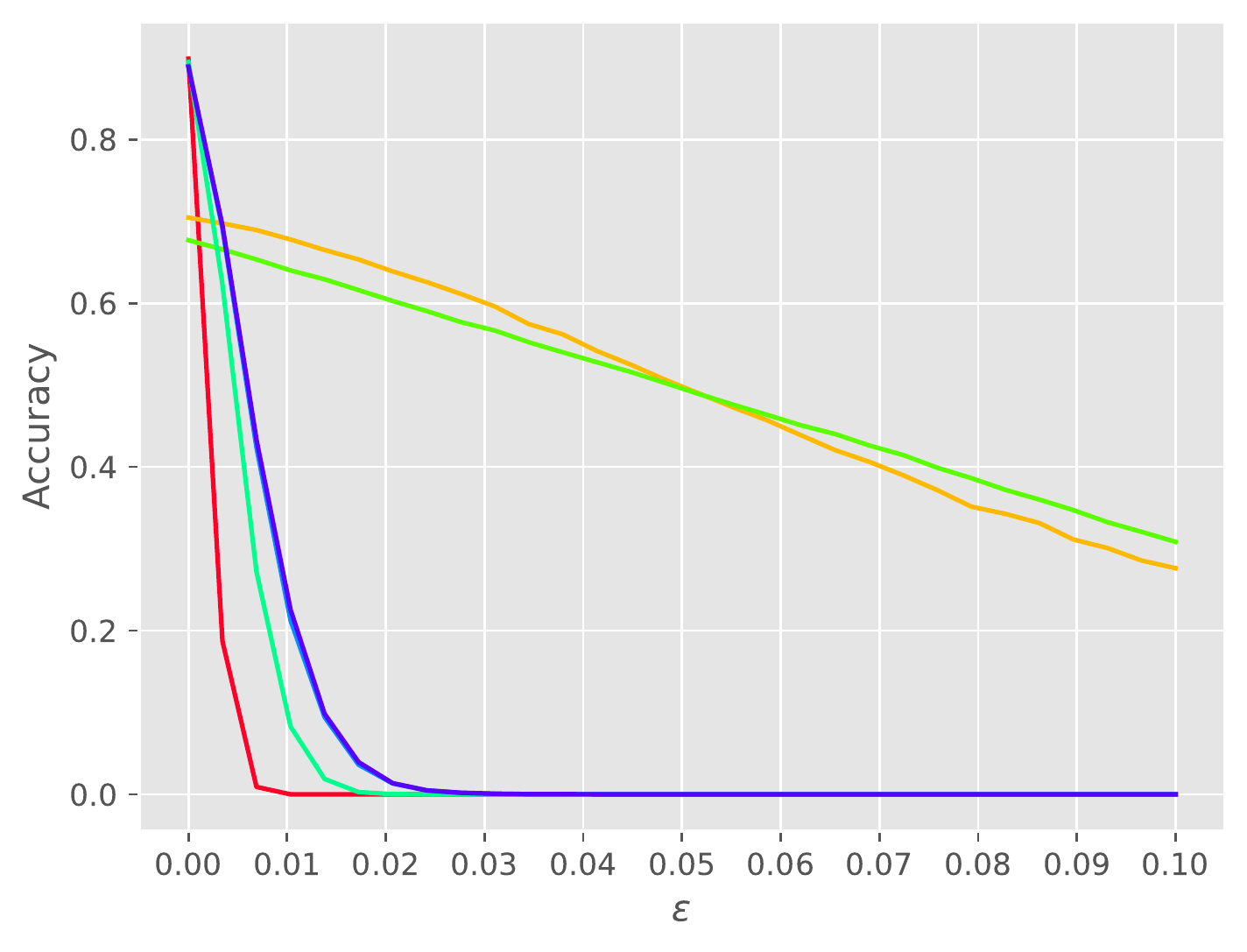}\label{fig:pgd-adam}}
    \subfigure[TPGD (Adam)]{\includegraphics[width=0.22\textwidth]{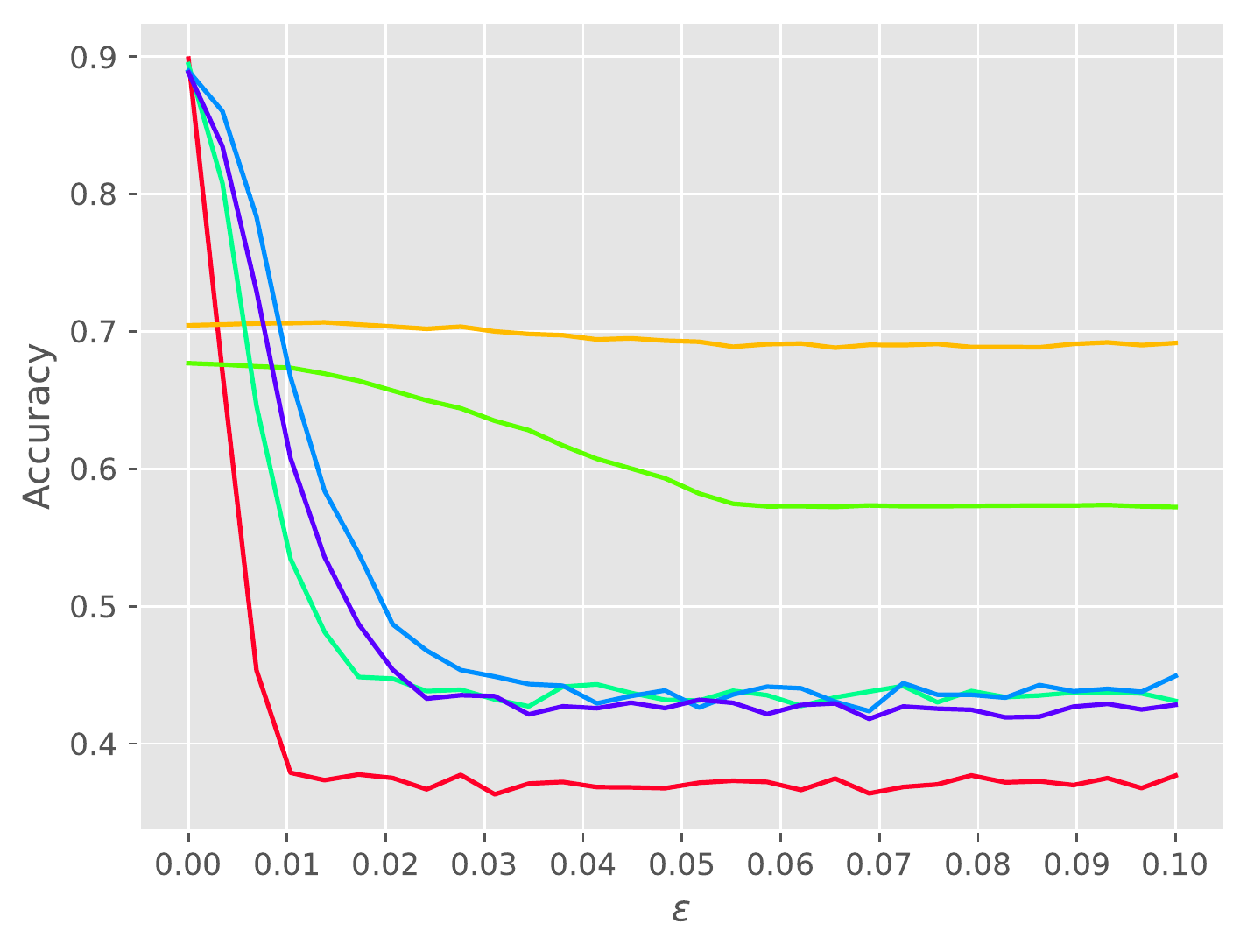}\label{fig:tpgd-adam}} 
    \caption{Accuracy of ResNet-20 on CIFAR-10 when trained with SGD or Adam under different white-box adversarial attacks: FGSM in (a) and (e), PGD in (b) and (f), APGD in (c) and (g), and TRADES attack in (d) and (h).}
    \label{fig:whitebox-attacks}
\end{figure}

\textbf{White-box attacks} ~~We vary the perturbation magnitude, $\epsilon$, from 0 to 0.1 with a step size of 0.01. We used the $L_{\infty}$ as the distance measure for PGD, APGD and TPGD. Fig.~\ref{fig:whitebox-attacks} depicts the ResNet-20 accuracy as function of $\epsilon$. There, when trained with FA and DFA, the model accuracy does not decrease significantly, unlike the models transporting either full weights or their sign only. We also notice that frSF and brSF, which transport the sign of the weight along with a random magnitude, deliver noisier gradients than uSF, thereby increasing the accuracy of the model as function of $\epsilon$. For both SGD and Adam optimizers, and in accordance with the accuracy vs. robustness trade-off \cite{zhang2019theoretically}, the accuracy of the models in Table ~\ref{tab:cifar10results} is complementary to their robustness.

\begin{figure}[h!]
    \centering
    \scriptsize{\includegraphics[scale=0.8]{images/robustness/whitebox/legend_algos.pdf}}\\
    \subfigure[Few-pixel (SGD)]{\includegraphics[width=0.22\textwidth]{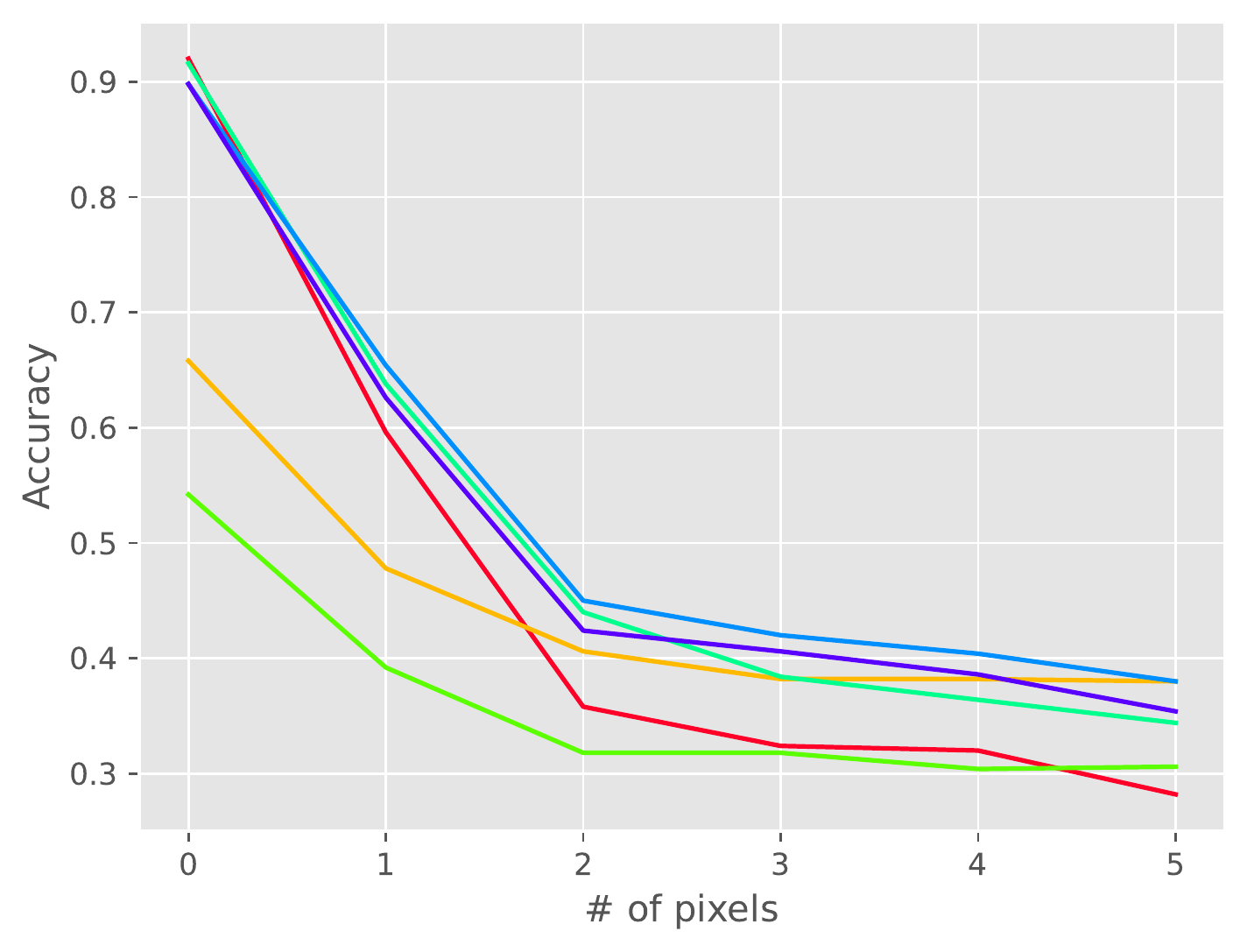}\label{fig:fewpixel-sgd}}
    \subfigure[Square (SGD)]{\includegraphics[width=0.22\textwidth]{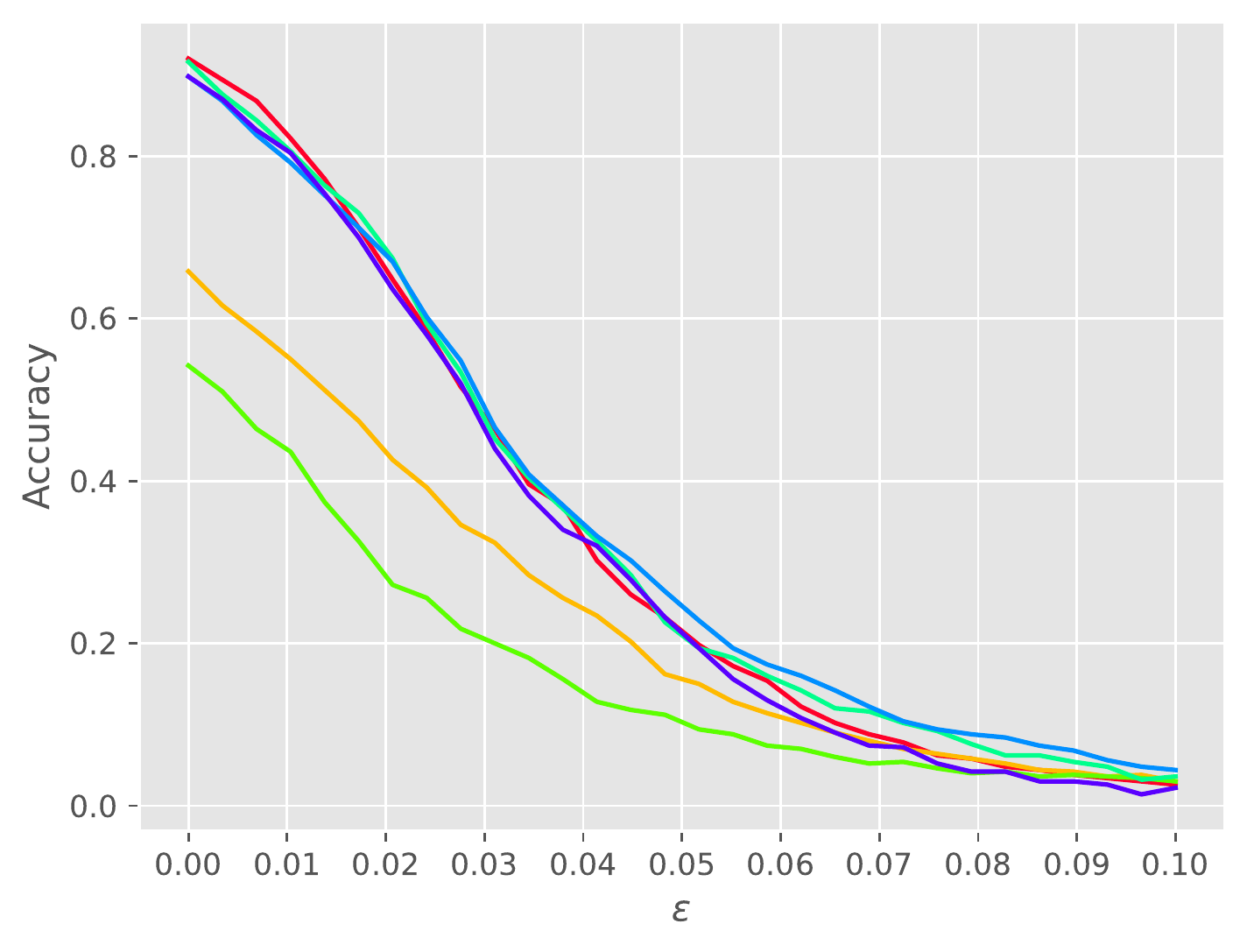}\label{fig:square-sgd}}\\
    \subfigure[Few-pixel (Adam)]{\includegraphics[width=0.22\textwidth]{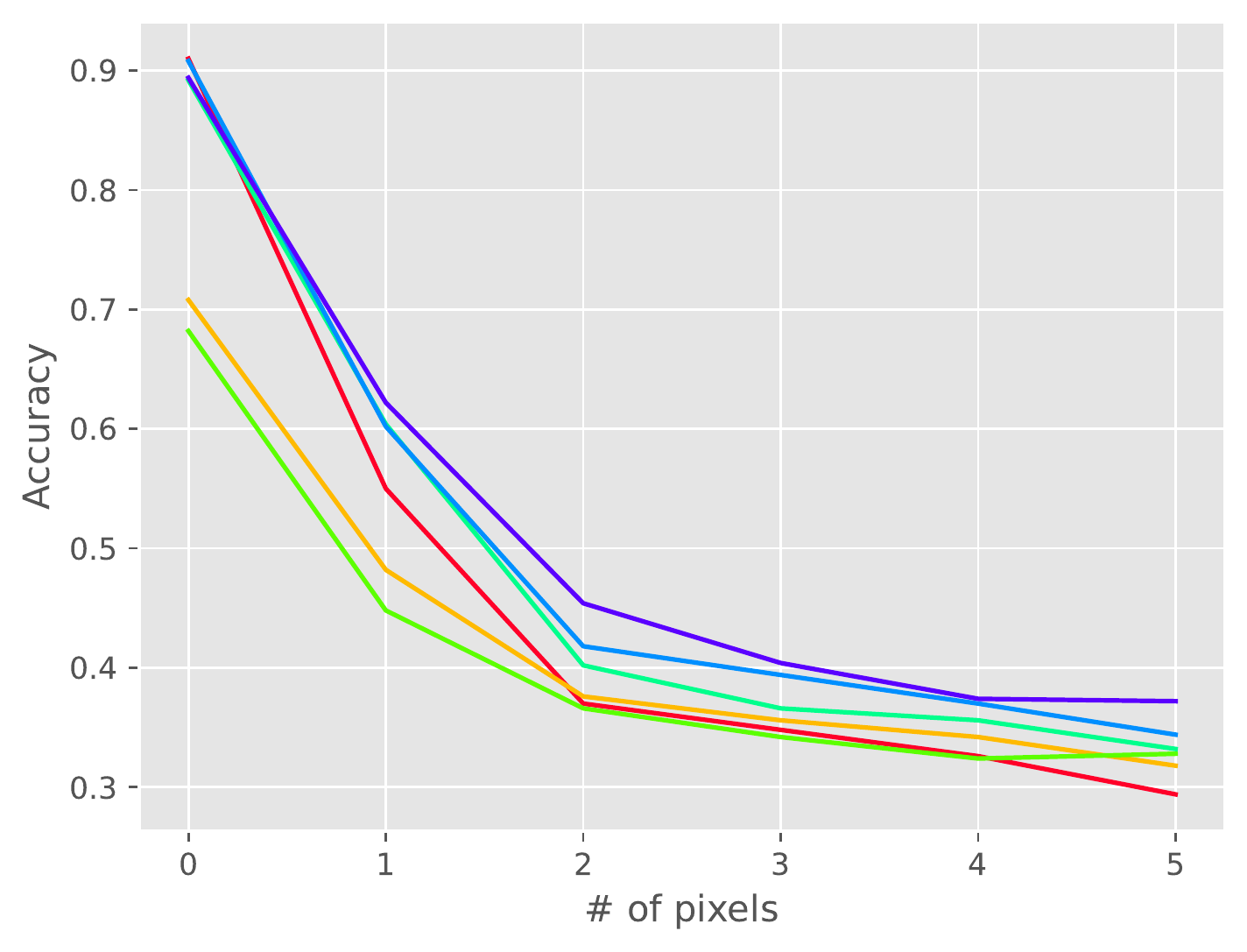}\label{fig:fewpixel-adam}} 
    \subfigure[Square (Adam)]{\includegraphics[width=0.22\textwidth]{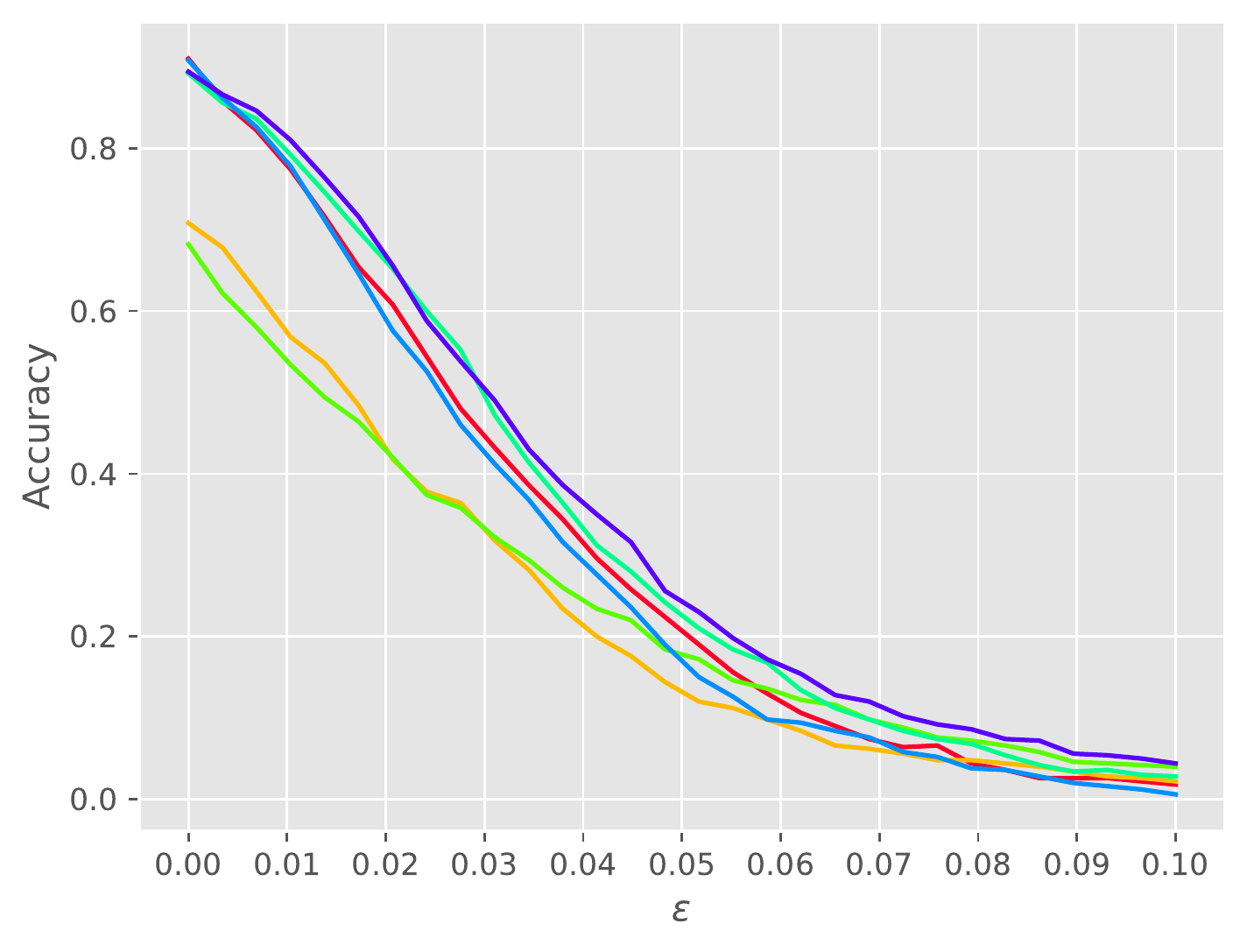}\label{fig:square-adam}}
    \caption{Accuracy of ResNet-20 on CIFAR-10 when trained with SGD or Adam under two black-box adversarial attacks: few-pixel in (a) and (c), and Square in (b) and (d).}
    \label{fig:blackbox-attacks}
\end{figure}

\textbf{Black-box attacks} ~~These family of attacks don’t intrinsically rely on gradients and only have access to the target model outputs. Figs.~\ref{fig:fewpixel-sgd} and \ref{fig:fewpixel-adam} show that the Few-Pixel attack is effective at finding adversarial examples as the number of function of the number of pixels allowed to change increases. FA, and sign concordant methods are more resistant to this attack than BP. However, the degradation of the model robustness under the Square attack, depicted in Fig.~\ref{fig:square-sgd} and \ref{fig:square-adam}, shows that the training of networks using synthetic gradient schemes did not provide a performance improvement against it.

\section{Conclusion}
We have presented \textit{BioTorch}, an open-source project that includes the implementation of feedback alignment methods and a framework for benchmarking them. To the best of our knowledge, this is the first framework that allows for effortless training and evaluation of neural networks using feedback alignment methods. Along with the description of \textit{BioTorch}'s main features, we have provided an accuracy and robustness benchmark of the methods mentioned in the Section~\ref{sec:alignment-methods_subsect} of this paper.

While \textit{BioTorch} is a step forward for the rigorous evaluation of biologically-motivated models, there are a number of extensions that could improve it, such as increasing support for other types of layers, including other existing biologically-motivated algorithms, or adding benchmarks for different domains and tasks beyond computer vision. We hope the release of our software will help the research community and encourage the design of new algorithms inspired by the human brain. 

{\small
\bibliographystyle{unsrt}
\bibliography{refs}
}

\clearpage
\section{Appendix}
\subsection{Experiment Details and Hyperparameters} \label{sec:experiment_details}
For all datasets, initial learning rates where selected by performing a grid search within the values [0.1, 0.01, 0.001, 0.0001].

\textbf{MNIST \& Fashion-MNIST}~~The modified LeNet network architecture chosen corresponds to the first column of Table~\ref{tab:networks-architecture}. Networks were trained with the Stochastic Gradient Descent (SGD) optimizer with a momentum of $0.9$, and weight decay of $10^{-3}$. Images were resized to $32 \times 32$ prior to being input to the network. We trained with a batch size of 64 for 100 epochs in one GPU. We decreased the initial learning rate by a factor of $10$ at the $50$th and the $75$th epoch. The initial learning rates chosen for every method can be found in Table~\ref{tab:hyperparams-lenet}. 

\begin{table}[!htbp]
\centering
\begin{tabular}{@{}ccc@{}}
\cmidrule{1-3}
LeNet layers               &  MNIST           &  CIFAR           \\ \midrule
Input                & 1$\times$32$\times$32               & 3$\times$32$\times$32               \\ \midrule
1                    & conv: 5$\times$5$\times$6, ReLU     & conv: 3x$\times$3$\times$32, ReLU    \\ \midrule
2                    & maxpool 2$\times$2 Stride: 2 & maxpool 2$\times$2 Stride: 2 \\ \midrule
3                    & conv: 5$\times$5$\times$16, ReLU    & conv: 3$\times$3$\times$64, ReLU    \\ \midrule
4                    & maxpool 2$\times$2 Stride: 2 & maxpool 2$\times$2 Stride: 2 \\ \midrule
5                    & conv: 5$\times$5$\times$120, ReLU   & conv: 3$\times$3$\times$128, ReLU   \\ \midrule
6                    & linear: 84, ReLU      & linear: 256, ReLU     \\ \midrule
7                    & linear: 10            & linear: 10            \\ \bottomrule
\end{tabular}
\vspace{0.2cm}
\caption{LeNet Architectures: the convolutional layers format is $K \times K \times C$,  where $K$ is the filter size, $C$ is the number of output channels. Stride is 1 by default if not further specified. The maxpool layer format is $K \times K$, where $K$ is the filter size. The linear format is $C$, where $C$ specifies the number of output neurons.}
\label{tab:networks-architecture}
\end{table}

\textbf{CIFAR10}~~The modified LeNet network architecture chosen corresponds to the second column of Table~\ref{tab:networks-architecture}. The ResNet-20 and ResNet-56 architectures \cite{he2016deep} have not been further modified. Networks using the SGD optimizer were trained with a momentum of $0.9$ and weight decay of $10^{-4}$. Networks using Adam were trained with the same weight decay and betas parameters equal to $[0.9, 0.999]$. We trained with a batch size of $128$ for $250$ epochs in one GPU. We decreased the initial learning rate by a factor of $10$ at the $100$th, $150$th and $200$th epoch. All the initial learning rates for both SGD and Adam can be found in Table~\ref{tab:hyperparams-lenet} for the LeNet architecture and in Tables~\ref{tab:hyperparams-resnet20} and \ref{tab:hyperparams-resnet56} for the ResNet ones. 

\begin{table}[!htbp]
\centering
\begin{tabular}{@{}cccccccc@{}}
\toprule
Dataset                  & Optimizer & BP    & FA    & DFA    & uSF   & brSF  & frSF     \\ \midrule
MNIST \& Fashion-MNIST   & SGD       & 0.001 & 0.001 & 0.0001 & 0.001 & 0.001 & 0.001   \\ \midrule
\multirow{2}{*}{CIFAR10} & SGD       & 0.1   & 0.001 & 0.0001 & 0.1   & 0.1   & 0.1    \\ \cmidrule(l){2-8} 
                         & Adam      & 0.001 & 0.0001 & 0.0001 & 0.001 & 0.001 & 0.001   \\ \bottomrule
\end{tabular}
\vspace{0.2cm}
\caption{Initial learning rates for LeNet style networks trained on MNIST, Fashion-MNIST and CIFAR-10, for SGD and Adam optimizers.}
\label{tab:hyperparams-lenet}
\end{table}

\begin{table}[!htbp]
\centering
\begin{tabular}{@{}ccccccc@{}}
\toprule
Optimizer     & BP    & FA  & DFA & uSF   & brSF  & frSF      \\ \midrule
SGD  & 0.1   & 0.1  & 0.001 & 0.1   & 0.1   & 0.1    \\ \midrule
Adam & 0.001 & 0.001 & 0.001 & 0.001 & 0.001 & 0.001  \\ \bottomrule
\end{tabular}
\vspace{1mm}
\caption{Initial learning rates for ResNet-20 trained on CIFAR-10, for the SGD and the Adam optimizers. }
\label{tab:hyperparams-resnet20}
\end{table}

\textbf{ImageNet}~~ A ResNet-18 network is trained with a batch size of $256$ and 2 GPUs for $75$ epochs using SGD with a initial learning rate of $0.1$. A scheduler decreased the learning rate by a factor of $10$ at the $20$th, the $40$th and the $60$th epoch. We used a weight decay of $10^{-4}$ and a momentum of $0.9$. For DFA we used Adam with an initial learning rate of $0.001$. At training time, a random resized crop of dimensions $224 \times 224$ of the original image or its horizontal flip with the per-pixel mean subtracted is used. When testing, the image is resized to $256 \times 256$ and then a center crop of $224 \times 224$ is used as input to the network.

\begin{table}[!htbp]
\centering
\begin{tabular}{@{}ccccccc@{}}
\toprule
Optimizer     & BP    & FA  & DFA  & uSF   & brSF  & frSF      \\ \midrule
SGD  & 0.1   & 0.01  & 0.0001 & 0.1   & 0.1   & 0.1    \\ \midrule
Adam & 0.001 & 0.001 & 0.001 & 0.001 & 0.001 & 0.001  \\ \bottomrule
\end{tabular}
\vspace{1mm}
\caption{Initial learning rates for ResNet-56 trained on CIFAR-10, for the SGD and the Adam optimizers.}
\label{tab:hyperparams-resnet56}
\end{table}

\subsection{BioTorch Configuration File}\label{appendix:conf-file}

Configuration files are structured in 6 parts as shown in the example in Listing.~\ref{ls:config_file_example}:

\textbf{Experiment} ~~This section includes general information about the experiment.

\begin{enumerate}[\hspace{0.5cm}]
    \item name (string): Name of the experiment.
    \item output\_dir (string):  Folder where the outputs of the experiment will be saved.
    \item seed (int): Random seed for the experiment.
    \item deterministic (boolean): If true, will apply \textit{PyTorch} deterministic operations to guarantee reproducibility.
\end{enumerate}

\textbf{Data} ~~This section includes information about the data that will be used to train the model.

\begin{enumerate}[\hspace{0.5cm}]

    \item dataset (string): Name of the dataset (see supported datasets in the code repository)
    \item dataset\_path (string): If dataset is not supported by \textit{Torchvision}, a path can be provided (e.g., ImageNet)
    \item target\_size (int): Images will be resized to the target size (if specified)
    \item num\_workers (int): Number of threads for data loading 
    
\end{enumerate}

\textbf{Model} ~~This section includes information about the model. For a list of supported architectures, initializations and feedback alignment methods we redirect the reader to the code repository. 

\begin{enumerate}[\hspace{0.5cm}]

    \item architecture (string): Name of the model architecture 
    \item mode/type (string): Name of the feedback alignment method 
    \item mode/options/init (string): Layer initialization method
    \item mode/options/gradient\_clip (boolean): If true, will apply gradient clipping (-1, 1)
    
\end{enumerate}

\textbf{Training Parameters} ~~This section includes information about the training hyperparameters. As it is very standard and for brevity, we redirect the reader to the configuration file example and the code repository to see the methods supported. We highlight the inclusion of the tracking alignment metrics. 

\begin{enumerate}[\hspace{0.5cm}]

    \item metrics/weight\_alignment (boolean): Track the angle between layers forward and backward weights
    \item metrics/weight\_ratio (boolean): Track the norm ratio between layers forward and backward weights
   
\end{enumerate}

\textbf{Infrastructure} This section includes information about the hardware. 
\begin{enumerate}[\hspace{0.5cm}]
    \item gpus/ (int/list): If -1 will use CPU, else will use the corresponding GPUs devices
\end{enumerate}

\textbf{Evaluation} This section includes information about the evaluation.
\begin{enumerate}[\hspace{0.5cm}]
    \item evaluation/ (boolean): If true, it will run an evaluation in the test and store the metrics after the training
\end{enumerate}

\begin{lstlisting}[language=yaml, label={ls:config_file_example}, basicstyle=\footnotesize, caption=BioTorch configuration file example to train a ResNet-20 FA model on CIFAR10.]
# 1. Experiment 
experiment:
  name: "fa_lr_0.1"
  output_dir: "trained_models_final/cifar10/resnet20/"
  # For reproducibility
  seed: 2021
  deterministic: true

# 2. Data
data:
  dataset: "cifar10_benchmark"
  dataset_path: null
  target_size: 32
  num_workers: 0

# 3. Model
model:
  architecture: "resnet20"
  mode:
    type: "fa"
    options:
      init: "xavier"
      gradient_clip: false

  pretrained: false
  # checkpoint:
  loss_function:
    name: "cross_entropy"

# 4. Training Parameters
training:
  hyperparameters:
    epochs: 250
    batch_size: 128

  optimizer:
    type: "SGD"
    lr: 0.1
    weight_decay: 0.0001
    momentum: 0.9

  lr_scheduler:
    type: "multistep_lr"
    gamma: 0.1
    milestones: [100, 150, 200]

  metrics:
    top_k: 5
    display_iterations: 500
    weight_alignment: true
    weight_ratio: true

# 5. Infrastructure
infrastructure:
  gpus: 0

# 6. Evaluation
evaluation: true
\end{lstlisting}

\subsection{Adversarial Attacks Description}\label{appendix:adversarial-attacks}
\textbf{FGSM}~~One of the earliest versions of to generate an adversarial example $x^{\textrm{adv}}$ using a model to attack $f$ is  the Fast Gradient Sign Method  (FGSM) \cite{goodfellow2014explaining} which updates the original image $x$, whose label is $y^*$, using the direction of the sign of the gradient with respect to the image as follows:
\begin{equation}
\label{eq:fgsm-attack}
x^{\textrm{adv}} = x + \epsilon \;\textrm{sign}\big(\nabla_x\,\mathcal{L}(f(x), y^*)\big)
\end{equation}

\textbf{PGD, APGD and TPGD}~~The Projected Gradient Descent (PGD) \cite{madry2017towards} is a variant of FGSM attack applied multiple times with a step size $\alpha$. That is to say:
\begin{equation}\label{eq:bmi-attack}
x^{\textrm{adv}}_0 = x,\quad\quad x^{\textrm{adv}}_{t+1} = \textrm{Clip}_{x}^{\epsilon}\bigg( x^{\textrm{adv}}_{t}  + \alpha \;\textrm{sign}\big(\nabla_{x^{\textrm{adv}}_{t}}\,\mathcal{L}(f(x^{\textrm{adv}}_{t}), y^*)\big)\bigg)
\end{equation}
where $x^{\textrm{adv}}_{t}$ denotes the adversarial example after $t$-steps and $\textrm{Clip}_{x}^{\epsilon}(\cdot)$ refers to the function that clips its parameter to be within the $\epsilon$-ball of center the original input $x$. The averaged PGD (APGD)\cite{zimmermann2019comment} estimates the gradient using the approximation $\frac{1}{k}\,\sum_{i}^{k}\nabla_{x^{\textrm{adv}}_{t}}\,\mathcal{L}(f(x^{\textrm{adv}}_{t}), y^*) \approx \mathbb{E}\left[\nabla_{x^{\textrm{adv}}_{t}}\,\mathcal{L}(f(x^{\textrm{adv}}_{t}), y^*)\right]$. The TPGD variant of PGD was used in the TRADES adersarial training \cite{zhang2019theoretically} with the only difference of the KL divergence loss instead of the cross-entropy loss function.

\textbf{Few-Pixel}~~Carefully designed perturbations that are even unrecognizable by human eyes can be appropriately perturbed using the one-pixel attack \cite{su2019one} based on differential evolution. While  common adversarial attacks are constructed by perturbating all pixels with an overall constraint on the strength
of the accumulated modification (e.g., the perturbation $\epsilon$ in FGSM and the ball radius $\epsilon$ and the step size $\alpha$ in PGD), the few-pixel attack limits the number of pixels to modify without confining the strength of modification. In this paper we consider the n-pixel attack where $\textrm{n}\in\{1,2,3,4,5\}$.

\textbf{Square:}~~Without the need for the local gradient information, the square attack \cite{andriushchenko2020square} is a score-based that is based on a randomized search scheme which selects localized square shaped updates at random positions to iteratively generate adversarial samples on the surface of the $l_{\infty}$- or $l2$-balls. This attack outperforms many more sophisticated attacks in terms of success rate and query efficiency.

\clearpage

\end{document}